\pdfoutput=1
\documentclass[11pt]{article}

\usepackage[utf8]{inputenc}
\usepackage{CJKutf8}
\usepackage[final]{acl}
\usepackage{arydshln}
\usepackage{times}
\usepackage{latexsym}

\usepackage[T1]{fontenc}
\usepackage[utf8]{inputenc}

\usepackage{microtype}

\usepackage{inconsolata}

\usepackage{tikz}
\usepackage{graphicx}
\usepackage{multirow}
\usepackage{amsmath}
\usepackage{amssymb}
\usepackage{wrapfig}   % 用于实现文字环绕图片
\usepackage{float}
\usepackage{algorithm}
\usepackage{algorithmic}
\usepackage{mdframed}
\usepackage{tcolorbox}
\usepackage{subcaption}
\usepackage{listings}
\usepackage{lipsum}
\usepackage{array}
\usepackage{chngcntr}
\usepackage{ragged2e}
\tcbuselibrary{breakable}
\usepackage{booktabs}
\usepackage{tabularx}
\usepackage{makecell}
\usepackage{caption}
\usepackage{xcolor} % Added for color definitions
\usepackage{pifont}  % For \ding{55} (a cross)
\usepackage{adjustbox}

\usepackage{booktabs}
\usepackage{multirow}
\usepackage[table]{xcolor}

\usepackage[table]{xcolor} % 用于表格颜色     % 用于单元格内换行
\definecolor{lightgray}{gray}{0.9} % 定义一个浅灰色
\usepackage{tcolorbox}
\tcbuselibrary{skins}
\usepackage{booktabs}
\usepackage{colortbl}
\usepackage[table]{xcolor}
\usepackage{siunitx}
\usepackage{xfp} % for \fpeval in \numcmp
\definecolor{best}{RGB}{255,235,190}   % light orange
\definecolor{second}{RGB}{220,235,255} % light blue

\usepackage{xstring}   

\definecolor{myGreen}{RGB}{0, 150, 0}
\definecolor{myRed}{RGB}{200, 0, 0}
\newcommand{\perf}[2]{%
    #1%
    \rlap{$
        \,_{\IfBeginWith{#2}{-}%
            {\color{myGreen}\text{\tiny{(#2)}}}%
            {\color{myRed}\text{\tiny{(#2)}}}%
        }
    $}%
}

\newtcolorbox{definitionbox}[1][]{%
  colback=blue!5,       % 背景浅蓝
  colframe=blue!50!black, % 边框颜色
  coltitle=white,       % 标题文字颜色
  colbacktitle=blue!60!black, % 标题背景色
  boxrule=1.5pt,                 % 边框粗细
  rounded corners,               % 设置为圆角
  fonttitle=\bfseries,  % 标题字体
  enhanced,
  attach boxed title to top left={yshift=-2mm,xshift=2mm},
  boxed title style={
    rounded corners,
    borderline west={0pt}{0pt}{white}, % 隐藏标题框自身的边框
    borderline east={0pt}{0pt}{white},
    borderline north={0pt}{0pt}{white},
    borderline south={0pt}{0pt}{white},
  },
  title=Definition,
  #1
}

\newcounter{assump}[section]

\newtcolorbox{assumbox}[2][]{%
  colback=blue!5,       % 背景浅蓝
  colframe=blue!50!black, % 边框颜色
  coltitle=white,       % 标题文字颜色
  colbacktitle=blue!60!black, % 标题背景色
  boxrule=1.5pt,                 % 边框粗细
  rounded corners,               % 设置为圆角
  fonttitle=\bfseries,  % 标题字体
  enhanced,
  breakable,
  attach boxed title to top left={yshift=-2mm,xshift=2mm},
  title={#2},
  boxed title style={
    rounded corners,
    borderline west={0pt}{0pt}{white}, % 隐藏标题框自身的边框
    borderline east={0pt}{0pt}{white},
    borderline north={0pt}{0pt}{white},
    borderline south={0pt}{0pt}{white},
  },
  before upper={\refstepcounter{assump}},
  #1
}

\title{How LoRA Remembers?\\ A Parametric Memory Law for LLM Finetuning}

\author{
Ziwen Xu\textsuperscript{1,2}\footnotemark[1],
~Haiwen Hong\textsuperscript{2}\footnotemark[1], 
~Linsong Yu\textsuperscript{1}\thanks{~~Equal Contribution.},
~Benglei Cui\textsuperscript{2},\\
~\textbf{Longtao Huang}\textsuperscript{2},
~\textbf{Hui Xue}\textsuperscript{2},
~\textbf{Ningyu Zhang}\textsuperscript{1}\thanks{~~Corresponding Author.}\\
\textsuperscript{1}Zhejiang University,
~\textsuperscript{2}Alibaba Group
}

\begin{document}
\maketitle

% 摘要和intro对应，三句话，第一句说大模型记忆（学习）新信息很重要。第二句，现有lora方法怎么怎么样不好。第三句，我们做了什么。
\begin{abstract}
Large Language Models (LLMs) must continuously learn and update knowledge to remain effective in dynamic real-world environments. While Low-Rank Adaptation (LoRA) is widely used for such memory updates, existing studies mainly rely on qualitative downstream evaluations, leaving the quantitative capacity limits and underlying dynamics of exact parametric memory largely unexplored. To bridge this gap, we employ LoRA as a controlled memory capacity probe within the latent space to systematically quantify exact parametric memory. We introduce the \textbf{Parametric Memory Law}, a robust power law linking loss reduction $\Delta\mathcal{L}$ to effective parameters and sequence length. At the token level, fine-grained analysis reveals a deterministic phase transition, demonstrating that a prediction probability of $p > 0.5$ constitutes a sufficient condition for verbatim recall under greedy decoding. Driven by these insights, we introduce \textbf{MemFT}, a threshold-guided optimization strategy that dynamically redistributes the training budget toward sub-threshold tokens. Empirical evaluations demonstrate that MemFT can enhance memory fidelity and efficiency\footnote{Code will be released at \url{https://github.com/zjunlp/ParametricMemoryLaw}.}.

% Large Language Models (LLMs) inherently store knowledge in static parameters, rendering the efficient encoding of post-deployment information under strict parameter budgets a fundamental challenge. 
% Using Low-Rank Adaptation (LoRA) as a memory capacity probe in the latent space, we systematically quantify the boundaries and dynamics of exact parametric memory. 
% We first formalize global capacity bounds via the \textbf{Parametric Memory Law}, a robust power law linking loss reduction $\Delta\mathcal{L}$ to effective parameters and sequence length. 
% Complementing this global view, fine-grained token-level analysis reveals a deterministic phase transition, demonstrating that a prediction probability of $p > 0.5$ constitutes a sufficient condition to guarantee verbatim recall under greedy decoding. 
% Driven by these insights, we introduce \textbf{MemFT}, a threshold-guided optimization strategy that dynamically redistributes the training budget toward sub-threshold tokens. 
% Empirical evaluations demonstrate that MemFT significantly enhances memory fidelity and parameter efficiency. 
% Ultimately, this work establishes a mechanistic foundation for parametric memory, bridging the gap between theoretical scaling limits and practical fine-tuning methods.
\end{abstract}

\section{Introduction}

Large Language Models (LLMs) have shown strong capabilities across diverse tasks and are now widely used in real-world systems~\cite{zhao2023survey}. 
However, their knowledge is encoded in static pretrained parameters and remains largely fixed after deployment. 
In practice, models continuously encounter new information such as updated facts, user preferences, and task-specific knowledge~\cite{DBLP:conf/emnlp/YaoWT0LDC023}. 
Efficiently integrating such information therefore becomes an key problem in continual learning and memory systems.

\begin{figure}[t]
\centering
\small
\includegraphics[width=1\linewidth]{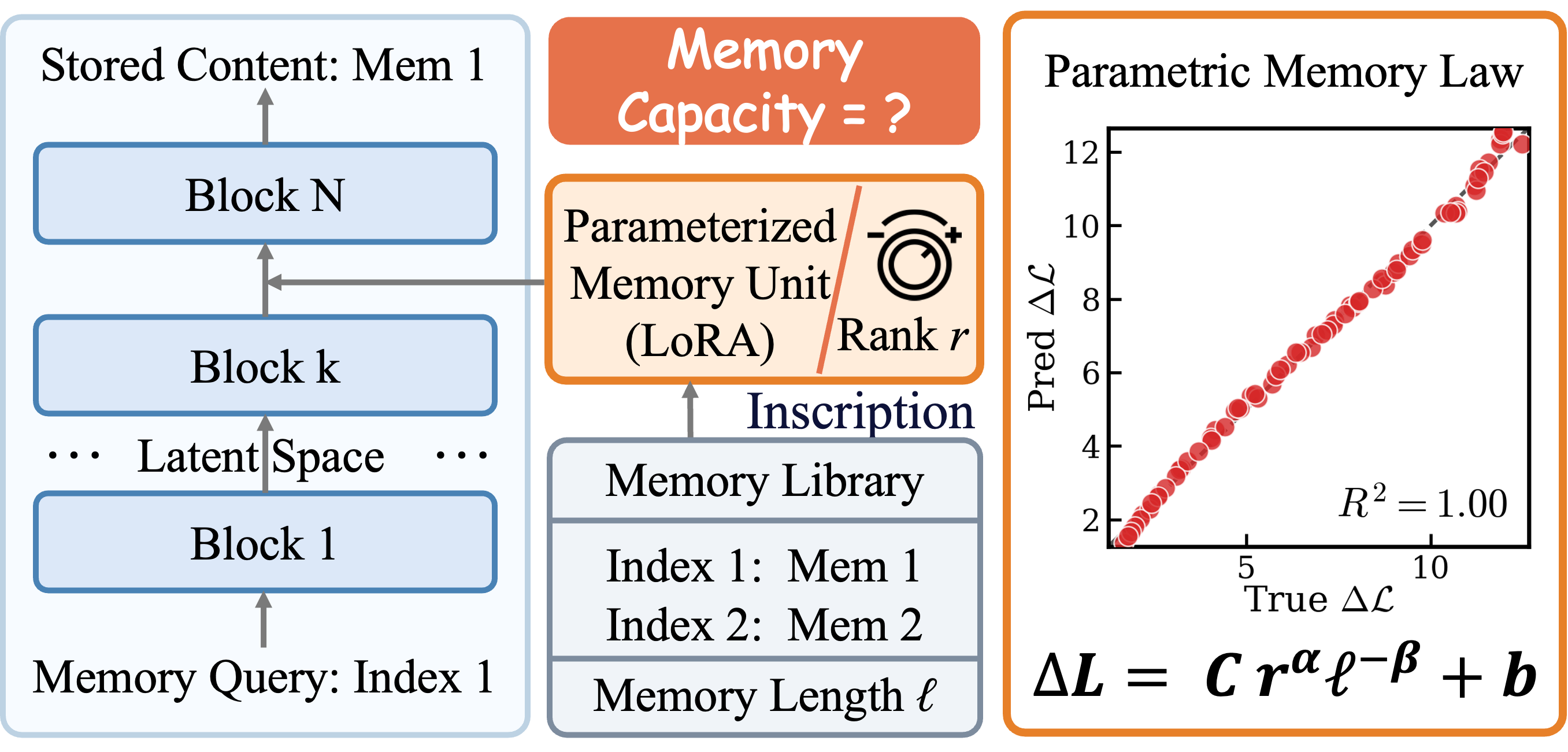}

\caption{
    LoRA as a pluggable memory unit in the LLM's latent space.
    The LoRA module (rank $r$) encodes contextual knowledge into the residual stream at layer $k$,
    enabling faithful recall of memorized text.
    The Parametric Memory Law quantifies the capacity-parameter trade-off.
}
\label{fig:motivation}
\vspace{-2em}

\end{figure}

Non-parametric methods address this challenge by providing external context during inference. 
Specifically, approaches such as in-context learning (ICL)~\cite{DBLP:conf/emnlp/Dong0DZMLXX0C0S24}, retrieval-augmented generation (RAG)~\cite{DBLP:journals/corr/abs-2312-10997}, and external non-parametric memory systems~\cite{DBLP:journals/corr/abs-2411-00489, DBLP:journals/corr/abs-2510-18866} dynamically integrate information without modifying model parameters. 
However, these methods are fundamentally constrained by fixed context windows, attention dilution, and escalating computational overhead as the input length scales~\cite{DBLP:journals/tacl/LiuLHPBPL24}.

In contrast, parametric memory embeds information directly into parameters or modular structures, enabling permanent knowledge consolidation and retrieval-free internal reasoning~\cite{DBLP:journals/corr/abs-2407-01178,DBLP:journals/corr/abs-2505-22101,Lei2026deltamemEO}. 
Recent works have further conceptualizes Low-Rank Adaptation (LoRA) as a specialized knowledge memory unit~\cite{DBLP:journals/corr/abs-2603-01097}. 
However, existing evaluations predominantly rely on downstream functional tasks, such as question answering. 
While effective for demonstrating practical utility of LoRA and its synergy with non-parametric methods~\cite{DBLP:journals/corr/abs-2603-01097,DBLP:conf/sigir/SuTAYWWY0025}, these benchmarks inevitably conflate raw information memorization with downstream comprehension and instruction-following behaviors. 
Consequently, the isolated quantitative capacity boundaries and dynamics mechanisms of parametric memory remain under-explored.

To bridge this gap, we focus on exact parametric memory. 
Drawing from fuzzy-trace theory in cognitive science~\cite{REYNA19951}, memory dual-encodes information into independent gist and verbatim traces. 
While functional benchmarks evaluate gist-level capability, exact text reconstruction isolates verbatim retention. 
Crucially, while non-parametric approaches guarantee verbatim output by directly fetching source text, parametric memory lacks this advantage and struggles with faithful reconstruction. 
Characterizing exact memory within parametric structures is therefore foundational. 
Using LoRA as a parameter-controllable probe within the latent space, we investigate:

\vspace{-0.4em}
\begin{assumbox}{Research Question}
What governing principles dictate the capacity boundaries and dynamic mechanisms of exact parametric memory?
\end{assumbox}
\vspace{-0.3em}

As illustrated in Figure~\ref{fig:motivation}, we investigate exact parametric memory by scanning rank and memory sequence configurations. 
Globally, the LoRA-induced loss reduction $\Delta\mathcal{L}$ follows a stable power-law scaling with effective parameters and sequence length, which we formalize as the \textbf{Parametric Memory Law}. 
At the token level, however, fine-grained analysis reveals demonstrate that a low average loss does not guarantee memorization.
Specifically, under greedy decoding, a token prediction probability of $p > 0.5$ is a sufficient condition to lock it into a stable state. 
Below this threshold, \textbf{stubborn tokens} face high-entropy competition with alternative tokens, sharply increasing the risk of autoregressive cascade failure. 
Based on these insights, we propose \textbf{MemFT}, an optimization strategy that redirects the parameter budget to sub-threshold tokens to maximize efficiency.

Our main contributions are:
\begin{itemize}
     \item \textbf{Parametric Memory Law:} We establish a power law that quantifies exact memory capacity based on parameters and sequence length.
    \item \textbf{Dynamics Mechanism:} We reveal that low average loss hides token-level competition, identifying $p > 0.5$ as a sufficient condition for memory locking and lower probabilities as a catalyst for cascade collapse.
    \item \textbf{MemFT Method:} We develop a threshold-guided optimization targeting stubborn tokens, significantly surpassing standard SFT in both fidelity and parameter efficiency.
\end{itemize}

\section{Preliminary}
\label{sec:preliminary}
% qoder生成参考内容，需要修改

\subsection{Exact Parametric Memory Task}
\label{sec:task}

\paragraph{Task Setup.}
Inspired by~\citet{DBLP:conf/icml/JelassiBKM24,DBLP:journals/corr/abs-2603-01097}, we formulate exact parametric memorization over a dataset $\mathcal{D} = \{(\mathbf{q}^{(i)}, \mathbf{a}^{(i)})\}_{i=1}^{N}$, where $\mathbf{q}^{(i)}$ serves as a unique \textbf{key} and $\mathbf{a}^{(i)} = (a^{(i)}_1, \ldots, a^{(i)}_{\ell_i})$ is the target content. 
Given a frozen base model $f_{\theta_0}$, we learn a parameter increment $\Delta\theta$ to construct an updated model $f_{\theta}$ with parameters $\theta = (\theta_0, \Delta\theta)$, satisfying:
\begin{equation}
    f_{\theta}(\mathbf{q}^{(i)}) = \mathbf{a}^{(i)}, \quad \forall i \in \{1, \ldots, N\}.
    \label{eq:success}
\end{equation}
Since $\mathbf{a}^{(i)}$ is inaccessible during inference except via the query $\mathbf{q}^{(i)}$, $\Delta\theta$ constitutes the exclusive medium for information storage. 
This reduces memorization to pure \emph{parameter writing}, decoupling it from retrieval or contextual comprehension.

\paragraph{Answer-only Accounting.}
Since questions serve only as keys, all token-level quantities in this paper (sequence length $\ell$, loss, accuracy) are computed exclusively over answer tokens; question tokens provide conditioning context and are excluded from every reported metric. 
Notably, the sequence length $\ell$ is determined by tokenizing the answer using each model's respective tokenizer.

\paragraph{Evaluation Metrics.}
Exact memorization demands a deterministic, reproducible decoding rule; thus, we adopt greedy decoding ($\hat{a}_t = \arg\max_{v\in\mathcal{V}} p_{\theta}(v \mid \mathbf{q}, a_{<t})$) throughout this work. 
We monitor three standard metrics to capture model behavior at different granularities.

Let $\mathbf{a} = (a_1, \dots, a_\ell)$ be the ground-truth answer of length $\ell$, and $\mathcal{L}_t(\theta) = -\log p_{\theta}(a_t \mid \mathbf{q}, a_{<t})$ be the token-level cross-entropy loss.

\textbf{Sequence-Averaged Loss.} 
The macroscopic loss serves as a global optimization proxy:
\begin{equation}
    \mathcal{L}(\mathbf{a}; \theta) = \frac{1}{\ell}\sum_{t=1}^{\ell} \mathcal{L}_t(\theta).
    \label{eq:avg_loss}
\end{equation}
Drawing on the view of language modeling as compression~\citep{DBLP:conf/iclr/DeletangRDCGMGW24}, we treat $\mathcal{L}(\mathbf{a}; \theta)$ as a measure of memorization for sequence $\mathbf{a}$, where the loss reduction $\Delta \mathcal{L}$ quantifies the memory gain.

\textbf{Token-Level Accuracy.} 
This metric measures the fraction of correctly predicted tokens, providing a microscopic view of memorization progress:
\begin{equation}
    \mathrm{Acc}_{\mathrm{tok}}(\mathbf{a}; \theta) = \frac{1}{\ell}\sum_{t=1}^{\ell} \mathbf{1}\!\left[\hat{a}_t = a_t\right].
    \label{eq:token_acc}
\end{equation}

\textbf{Exact Match Accuracy.} 
This strict binary metric evaluates whether the entire sequence is reproduced verbatim:
\begin{equation}
    \mathrm{Acc}_{\mathrm{EM}}(\mathbf{a}; \theta) = \mathbf{1}\!\left[ \forall t \in \{1,\dots,\ell\}, \hat{a}_t = a_t \right].
    \label{eq:em_acc}
\end{equation}

We observe that $\mathcal{L}$ and $\mathrm{Acc}$ are \emph{not} monotonically aligned. 
Consequently, we track all three metrics: $\mathcal{L}$ for global convergence trends, $\mathrm{Acc}_{\mathrm{tok}}$ for fine-grained token-level dynamics, and $\mathrm{Acc}_{\mathrm{EM}}$ for strict recall fidelity.

\subsection{LoRA-based Parametric Memory Injection}
\label{sec:lora}

We realize $\Delta\theta$ with LoRA: for each adapted linear layer with frozen weight $W_0 \in \mathbb{R}^{d_{\text{out}} \times d_{\text{in}}}$, we attach a low-rank residual branch so that its forward pass becomes
\begin{equation}
    h \;=\; W_0 x + B A x,
    A \in \mathbb{R}^{r \times d_{\text{in}}},\; B \in \mathbb{R}^{d_{\text{out}} \times r},
    \label{eq:lora}
\end{equation}
where $r \ll \min(d_{\text{in}}, d_{\text{out}})$ is the LoRA rank and $\Delta\theta$ collects all such $\{(A_\ell, B_\ell)\}_\ell$.
We view LoRA as a \emph{latent-space probe}: $r$ is a single monotone knob on the trainable parameter count, letting us sweep the capacity axis and cleanly examine how parameter budget relates to memory capacity.
Since $\theta_0$ is frozen, any change in $\mathcal{L}$ or $\mathrm{Acc}$ is attributable solely to $\Delta\theta$.
At inference time $\Delta\theta$ is used through the residual branch in Eq.~\ref{eq:lora}; whether $BA$ is fused back into $W_0$ is an implementation choice that leaves $f_\theta$ unchanged.

% 现象展示：绘制log(Rank)、log(Length) 与log(Loss) 的三维散点图。
% 核心发现：最低loss经过0.69的阈值筛选后，数据点在空间中呈现显著的平面线性分布。这表明 Loss 与 Rank、Length 之间存在潜在的幂律关系（Power-Law Relationship），而非简单的线性或多项式关系。
% 在本节中，我们通过大规模的定量实验，揭示了大语言模型（LLM）中参数化记忆的宏观缩放规律。我们首先展示了在双对数空间（Log-Log Space）中观察到的显著线性现象，随后基于此提出了参数化记忆定律（Parametric Memory Law），最后通过多数据集的拟合验证了该定律的普适性与鲁棒性。
% 现象展示：绘制log(Rank)、log(Length) 分别与log(Loss) 的二维图。
% 核心发现：最低loss经过0.69的阈值筛选后，数据点在log-log空间中中呈现显著的线性分布。这表明 Loss 与 Rank、Length 之间存在潜在的幂律关系（Power-Law Relationship），而非简单的线性或多项式关系。

% 参考论文：There Will Be a Scientific Theory of Deep Learning

\section{The Parametric Memory Law}
\begin{figure*}[t]
    \centering
    
    % --- Subfigure (a) ---
    \begin{subfigure}[t]{\textwidth}
        \centering
        \includegraphics[height=4cm]{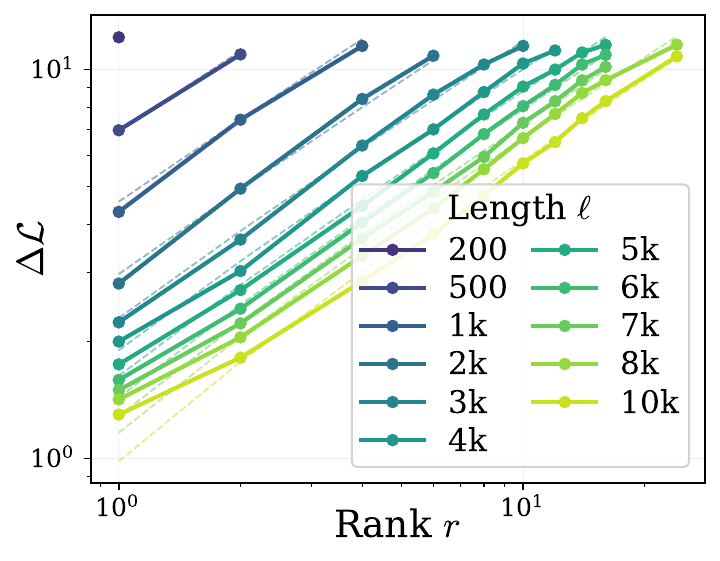}
        \hfill
        \includegraphics[height=4cm]{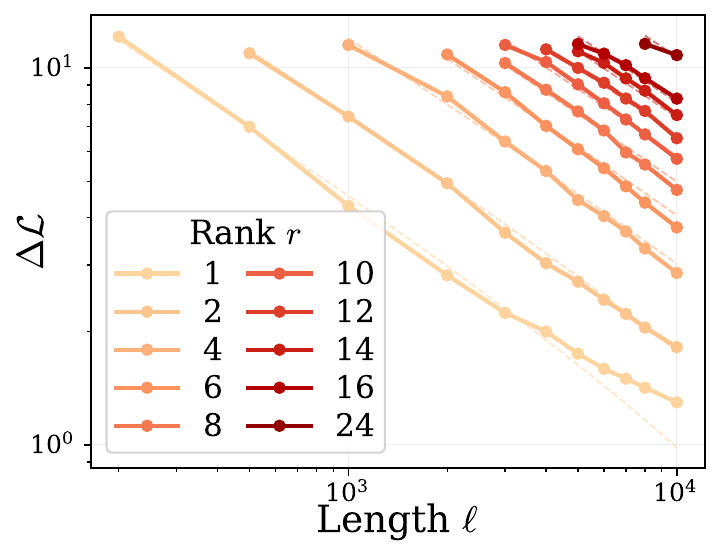}
        \hfill
        \includegraphics[trim=0 20 0 30, clip, height=4.3cm]{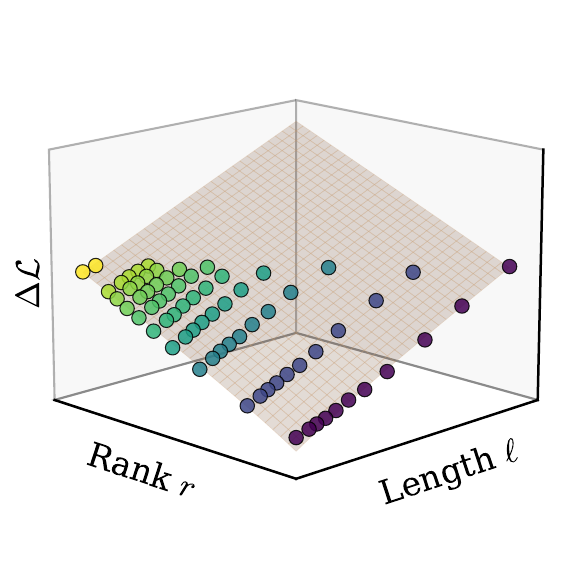}
        \caption{Approximate log-linear dependence of $\Delta\mathcal{L}$ on rank $r$ and length $\ell$}
        \label{fig:main-a}
    \end{subfigure}
    
    \vspace{6pt}
    
    % --- Subfigure (b) ---
    \begin{subfigure}[b]{0.30\textwidth}
        \centering
        \includegraphics[height=4.0cm]{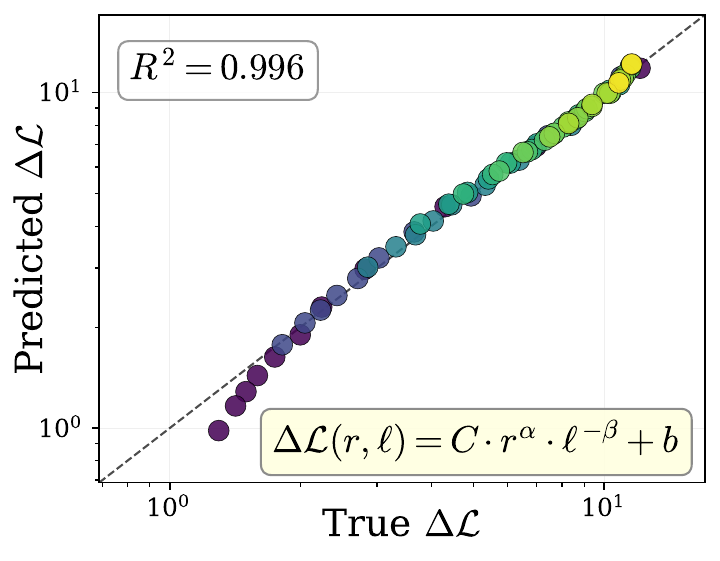}
        \caption{Predicted vs. true $\Delta\mathcal{L}$}
        \label{fig:main-b}
    \end{subfigure}
    \hfill
    \begin{subfigure}[b]{0.66\textwidth}
        \centering
        \includegraphics[height=4.0cm]{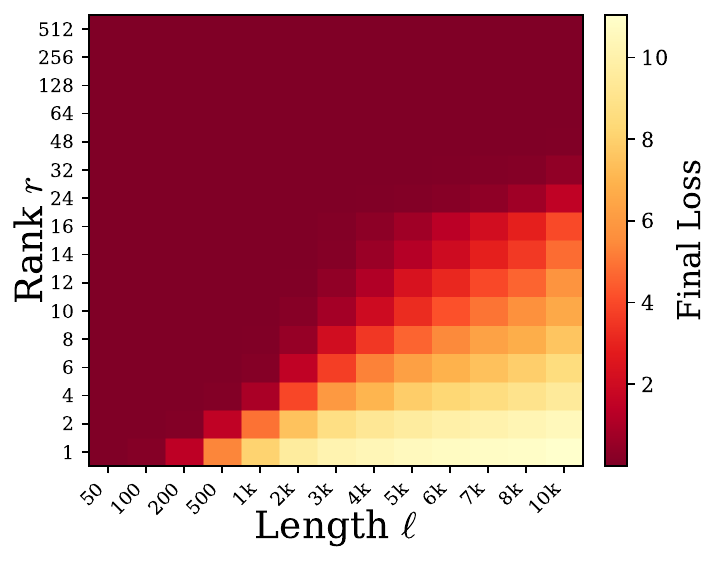}
        \hfill
        \includegraphics[height=4.0cm]{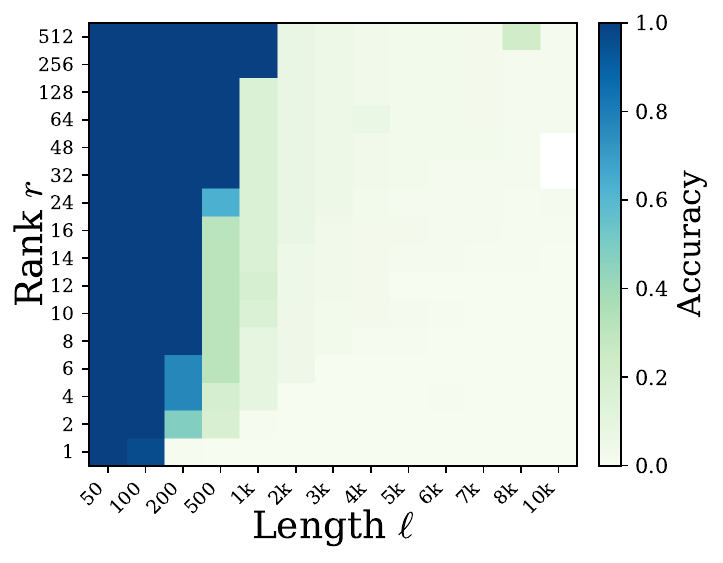}
        \caption{Decoupling of low loss and high accuracy}
        \label{fig:main-c}
    \end{subfigure}

    \caption{\textbf{Empirical validation of the Parametric Memory Capacity Law (LoRA on Qwen3-8B).} 
    \textbf{(a)} $\Delta\mathcal{L}$ exhibits approximate log-linear decay with respect to rank $r$ and length $\ell$, forming a nearly planar structure in log-space; 
    \textbf{(b)} The scatter plot compares predicted $\Delta\mathcal{L}$ from Eq.~\eqref{eq:parametric_law} against true values, showing high fidelity ($R^2=0.996$); 
    \textbf{(c)} Heatmaps plot the final loss and token-level accuracy (correct tokens / total length) across various $(r, \ell)$ settings, revealing numerous cases where loss approaches zero while accuracy remains near zero.}
    \label{fig:main}
    \vspace{-1.5em}
    
\end{figure*}

In this section, we discover the \textit{Parametric Memory Law} through large-scale quantitative experiments, which governs the macroscopic scaling behavior of parametric memory in LLMs.

\subsection{Empirical Observation: Linearity in Log-Log Space}
\label{sec:empirical_obs}
To investigate the scaling dynamics of parametric memory, we define \textbf{Loss Reduction} as $\Delta \mathcal{L} = \mathcal{L}_{init} - \mathcal{L}_{final}$, where $\mathcal{L}_{init}$ and $\mathcal{L}_{final}$ denote the cross-entropy losses before and after applying parametric memory, respectively.

We conducted experiments on Qwen3-8B-IT~\cite{DBLP:journals/corr/abs-2505-09388} and Llama3.1-8B-IT~\cite{DBLP:journals/corr/abs-2407-21783}.
The experimental design covered two typical scenarios: 
(1) a \textbf{Long-Context Memorization Stress Test}, inspired by~\cite{DBLP:conf/acl/ZhuLPJKL24}, using a LongBench~\cite{DBLP:conf/acl/BaiLZL0HDLZHDTL24} sample with 0\%-100\% token replacement by randomly sampled Qwen vocabulary tokens to generate different levels of semantic coherence and difficulty; 
(2) a \textbf{Short-Context Dense Memory Test} using PhoneBook datasets~\cite{DBLP:conf/icml/JelassiBKM24,DBLP:journals/corr/abs-2603-01097} to evaluate high-density storage limits.
We varied LoRA ranks $r$ and sequence lengths $\ell$ extensively across these settings.

We analyze the relationship among $\Delta \mathcal{L}$, $r$, and $\ell$ across a wide range of experimental settings.
As illustrated in Figure~\ref{fig:main-a}, we observe distinct linear trends in the log-log domain.
Specifically, $\Delta \mathcal{L}$ scales positively with rank $r$ and negatively with length $\ell$.
This high degree of linearity strongly suggests an underlying \textbf{power-law relationship} between $\Delta \mathcal{L}$ and the parameters $(r, \ell)$.

Empirically, we exclude saturated samples with $\mathcal{L}_{final} \le 0.69$, with the threshold's origin detailed in Section \ref{sec:deterministic_phase}.

\subsection{Formulating the Parametric Memory Law}

Based on the observed log-linearities, we formalize the scaling behavior as the \textbf{Parametric Memory Law} and propose the following empirical model:
\begin{equation}
    \Delta \mathcal{L}(r, \ell) = C \cdot r^{\alpha} \cdot \ell^{-\beta} + b
    \label{eq:parametric_law}
\end{equation}
Here, $\Delta \mathcal{L}$ denotes the training memory gain, while $C$ is a scaling constant dictated by model and data distribution.
$\alpha$ (\textbf{Capacity Exponent}) quantifies the efficiency of parameter rank in enhancing memory capacity. 
$\beta$ (\textbf{Length Penalty Exponent}) reflects the nonlinear increase in memory difficulty associated with longer sequences. 
$C, \alpha, \beta$ are positive.
% $b$ is an offset term accounting for baseline noise.

This law indicates that within the significant memory gain regime, performance is governed by a power-law trade-off between rank and length.
% ($\mathcal{L}_{final} > 0.69$)
% Notably, the threshold $0.69$ serves as an empirical boundary here; its microscopic origin is elucidated in Section \ref{sec:deterministic_phase}.

\subsection{Fitting Validation}
\label{sec:fitting_validation}

We validated the Parametric Memory Law (Eq.~\ref{eq:parametric_law}) against the experimental data from Section~\ref{sec:empirical_obs}, reporting both the coefficient of determination ($R^2$) and Mean Absolute Percentage Error (MAPE) to assess goodness-of-fit. 

As shown in Table~\ref{tab:fit_r2}, the law demonstrates \textbf{exceptional explanatory power across diverse models and data distributions}. 
Specifically, it achieves \textbf{$R^2 > 0.98$} with low MAPE in all settings, including pure semantic, fully random, and short-context PhoneBook tasks. 
Notably, the law exhibits \textbf{strong robustness to varying semantic densities}. 
A single unified formula accurately fits the entire Long-context mixture (0\%--100\% random), yielding high $R^2$ values of \textbf{0.987} for Llama3.1-8B-IT and \textbf{0.983} for Qwen3-8B-IT. 
These results confirm that the power law \textbf{precisely characterizes the scaling of loss reduction}. 
This consistency holds regardless of whether the context is structured or random, spanning from long-context to short-context scenarios.

In summary, the Parametric Memory Law provides a \textbf{robust macroscopic mapping between parameter budget, sequence length, and loss reduction}. 
However, by focusing on aggregate metrics, it abstracts away the microscopic dynamics of individual token memorization, which we analyze in the next section.

% Auto-generated by make_fit_table.py
% R^2 (linear space) of the parametric-memory law: delta_loss = C * r^alpha * L^beta + b

\begin{table*}[t]
\centering
% \small
\setlength{\tabcolsep}{5pt}
\begin{tabular}{llccccccc|c}
\toprule
\multirow{2}{*}{Model} & \multirow{2}{*}{Metric} & \multicolumn{7}{c}{Long-context memorisation stress test} & Phonebook \\
\cmidrule(lr){3-9}
 & & r0 & r20 & r40 & r60 & r80 & r100 & \textbf{Comb.} & All \\
\midrule
\multirow{2}{*}{Llama3.1-8B-IT} & $R^2$\,$\uparrow$ & 0.992 & 0.994 & 0.996 & 0.995 & 0.996 & 0.996 & \textbf{0.987} & 0.981 \\
 & MAPE\,(\%)\,$\downarrow$ & 1.430 & 2.493 & 2.528 & 2.755 & 2.710 & 2.563 & \textbf{7.057} & 1.606 \\
\midrule
\multirow{2}{*}{Qwen3-8B-IT} & $R^2$\,$\uparrow$ & 0.996 & 0.993 & 0.996 & 0.996 & 0.995 & 0.996 & \textbf{0.983} & 0.990 \\
 & MAPE\,(\%)\,$\downarrow$ & 0.752 & 2.553 & 2.331 & 2.862 & 3.944 & 3.472 & \textbf{8.320} & 0.476 \\
\bottomrule
\end{tabular}
\caption{Goodness-of-fit of the parametric-memory law $\Delta\mathcal{L}(r,\ell)=C\cdot r^{\alpha}\cdot\ell^{-\beta}+b$ (Eq.~\ref{eq:parametric_law}) on two parametric-memorization benchmarks. We report two complementary metrics: $R^2$ (higher is better) and MAPE (lower is better). The long-context memorization stress test sweeps the random-token ratio (r$x$$=$$x\%$ random; r0$=$pure LongBench, r100$=$fully random), and \textbf{Comb.}\ pools the six mixtures. Phonebook stores many short (name$\rightarrow$number) key--value pairs, probing the short-text memory regime complementary to long-context.}
\label{tab:fit_r2}
\vspace{-0.5em}

\end{table*}

\section{The Deterministic Phase Transition of Memory}
The parametric memory law describes the macroscopic scaling behavior, but the average loss metric masks the discrete nature of token-level memory.
This section reveals the misalignment between loss and accuracy and establishes the critical point of the deterministic phase transition that determines the success or failure of memory.

\subsection{The Loss-Accuracy Misalignment}
\label{sec:misalignment}

In exact parametric memory tasks, minimizing average cross-entropy loss does not guarantee high token-level accuracy, a phenomenon we term the \textbf{Loss-Accuracy Misalignment}.

Figure~\ref{fig:main-c} shows models achieving near-zero average loss yet negligible accuracy. This occurs because average loss smooths over local variations, allowing high confidence on easy tokens to mask catastrophic errors on hard ones. As illustrated in Figure~\ref{fig:stubborn-a}, specific positions maintain persistently low probabilities ($p < 0.5$) despite low global loss, creating invisible bottlenecks.

In autoregressive generation, such local errors are fatal. A single misprediction alters the context for subsequent steps, triggering error propagation that collapses the sequence. Thus, average loss is an insufficient proxy for generation fidelity, necessitating a shift to token-level probability analysis.

\subsection{Token-Level Probability Dynamics}
\label{sec:token_dynamics}

\begin{figure*}[t]
    \centering
    \begin{subfigure}[t]{0.32\linewidth}
        \centering
        \includegraphics[width=\linewidth]{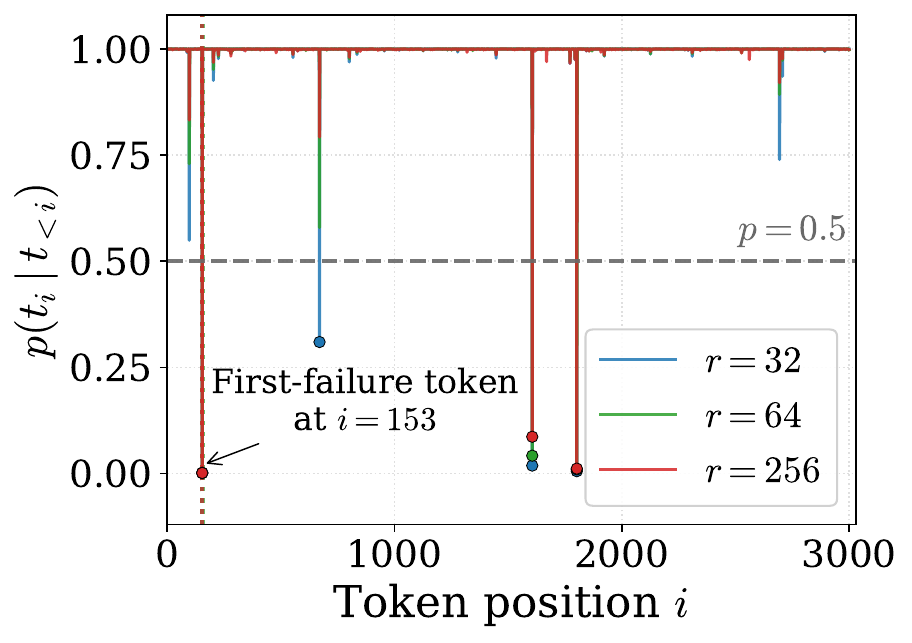}
        \caption{Probability dynamics across ranks}
        \label{fig:stubborn-a}
    \end{subfigure}\hfill
    \begin{subfigure}[t]{0.32\linewidth}
        \centering
        \includegraphics[width=\linewidth]{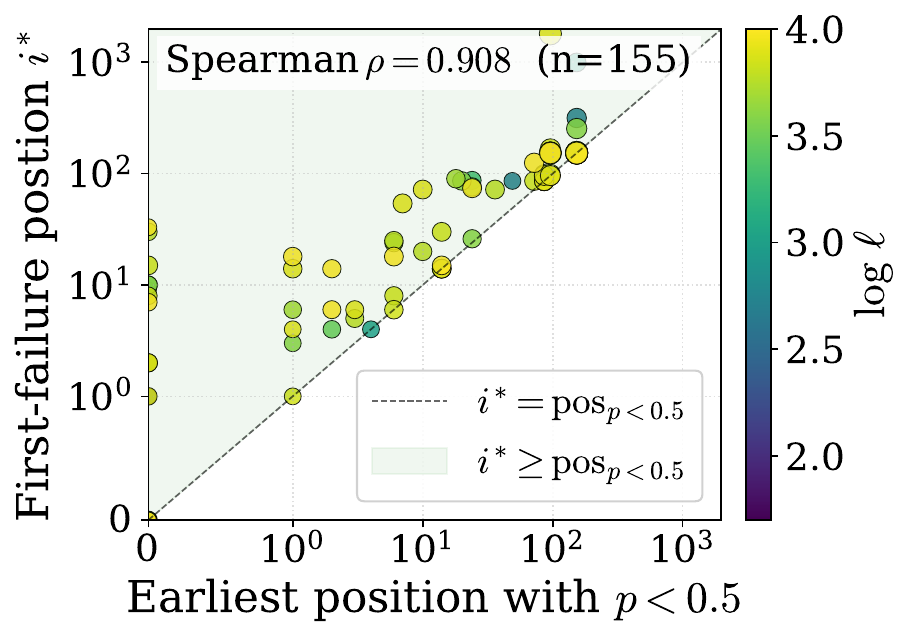}
        \caption{Lower bound on first failure}
        \label{fig:stubborn-b}
    \end{subfigure}\hfill
    \begin{subfigure}[t]{0.32\linewidth}
        \centering
        \includegraphics[width=\linewidth]{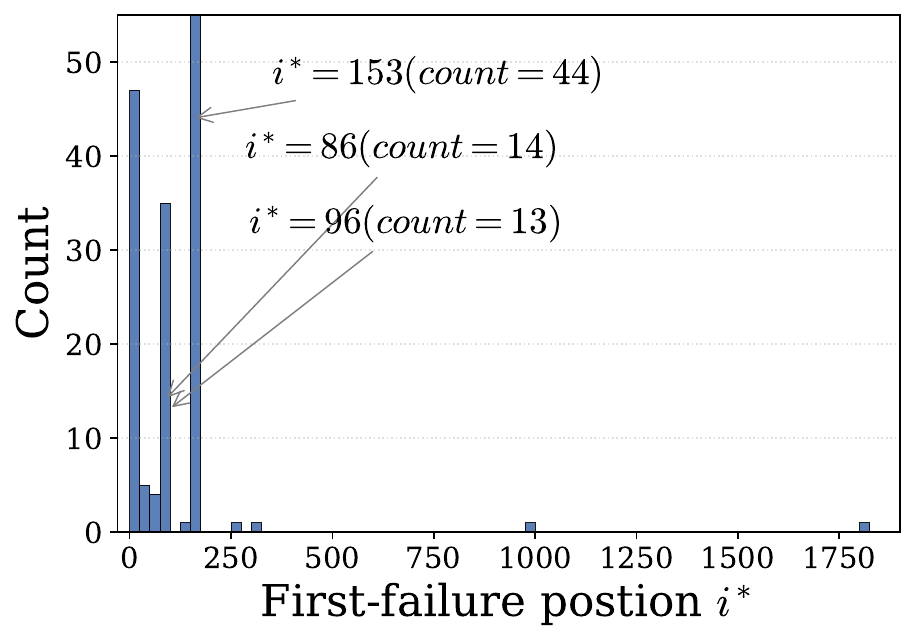}
        \caption{Localization of failure positions}
        \label{fig:stubborn-c}
    \end{subfigure}
    \caption{\textbf{Microscopic origin of the Loss-Accuracy Paradox.} 
    Results are based on Qwen3-8B trained on the Random dataset. 
    \textbf{(a)} \textbf{Sparse stubborn positions}: A small set of indices where target probabilities persistently remain $p<0.5$, resisting improvement even as LoRA rank increases. 
    \textbf{(b)} \textbf{Correlation with decoding failure}: The earliest stubborn position tightly bounds the first failure index $i^{\ast}$ (Spearman $\rho=0.908$, $n=155$), indicating that sub-threshold probabilities \textbf{significantly increase the likelihood} of errors due to lost probabilistic dominance. 
    \textbf{(c)} \textbf{Spatial concentration}: The histogram of first-failure positions across all settings reveals high localization.}
    \label{fig:stubborn}
    % \vspace{-1em}
\end{figure*}

To uncover the microscopic origin of the Loss-Accuracy Misalignment, we analyze the per-token probabilities $p(t_i \mid t_{<i})$ after SFT convergence.

We identify \textbf{stubborn token positions} as indices where target token probabilities remain persistently below the $p=0.5$ threshold, regardless of LoRA rank increases (Figure~\ref{fig:stubborn-a}; full grids across data scenarios in Appendix~\ref{app:stubborn_grid}).
These bottlenecks are highly localized; 
for instance, Figure~\ref{fig:stubborn-c} shows that position $i=153$ alone accounts for 28\% of all failures, indicating that these are intrinsic hard cases resistant to capacity scaling.

Crucially, these stubborn positions drive autoregressive collapse. 
As demonstrated in Figure~\ref{fig:stubborn-b}, the earliest stubborn position tightly bounds the first decoding failure $i^{\ast}$ (Spearman $\rho=0.908$). 
When $p < 0.5$, the correct token \textbf{loses probabilistic dominance} and becomes highly susceptible to being superseded by incorrect candidates during greedy decoding. 
This triggers cascading failures, \textbf{corrupting all subsequent tokens} and explaining why a single local bottleneck leads to complete sequence failure.

\subsection{Deterministic Phase Transition}
\label{sec:deterministic_phase}
% 说明相变临界点：P=0.5 是记忆状态的二元分界线。在贪婪解码设定下只要正确token概率高于0.5，即loss小于0.69，即可记忆正确。然后引出在记忆任务设定下loss降到0的必要性讨论。
The analysis above directs our attention to the critical role of the $p=0.5$ threshold. Under greedy decoding, this probability value serves as the boundary for deterministic memory success, leading us to define the \textbf{Deterministic Phase Transition}.

Greedy decoding selects the token with the highest predicted probability. 
For successful memory, the target token must be the most probable candidate. 
A sufficient condition to guarantee this dominance is $P_{\text{target}} > 0.5$, as no other single candidate can exceed this value if the target holds the majority of the probability mass. 
Thus, $P_{\text{target}} = 0.5$ acts as the critical threshold for deterministic success.

This probability threshold corresponds to a critical loss value. 
Given the cross-entropy loss $\mathcal{L} = -\log(P_{\text{target}})$, substituting $P_{\text{target}} = 0.5$ yields:
\begin{equation}
    \mathcal{L}_{\text{crit}} = -\log(0.5) = \ln(2) \approx 0.693
\end{equation}

This derivation provides the theoretical basis for the empirical threshold $\mathcal{L}_{final} > 0.69$ in Section \ref{sec:empirical_obs}.
We characterize the memory states relative to this boundary:
(1) \textbf{Disordered Phase} ($\mathcal{L} > \mathcal{L}_{\text{crit}}$): Here, $P_{target} < 0.5$. The correct token does not hold a dominant probability, making it susceptible to being outcompeted by other candidates, thus leading to potential memory failure.
(2) \textbf{Ordered Phase} ($\mathcal{L} < \mathcal{L}_{\text{crit}}$): Here, $P_{target} > 0.5$. The correct token is guaranteed to be the most probable candidate, ensuring successful reproduction under greedy decoding.

Thus, $\mathcal{L}_{\text{crit}}$ represents a sharp phase transition boundary between uncertain and deterministic memory success.
The Parametric Memory Law describes the scaling trend of loss reduction, while the Deterministic Phase Transition explains why loss must cross this barrier to translate into effective accuracy.
Pursuing lower loss aims to increase the confidence margin, but the acquisition of reliable memory capability begins with crossing this deterministic phase transition.

\section{MemFT: Methodology and Empirical Verification}
\label{sec:method}
% Auto-generated by make_main_table.py
% Metric: per-sample acc = correct/length (random has 1 sample per cell),
% averaged across all sequence lengths.

\begin{table*}[t]
\centering
% \small
% \setlength{\tabcolsep}{2.0pt} % 适中列间距

% 注意：原表有18列，去掉Model列后剩17列 (1个l + 16个c)
\resizebox{\textwidth}{!}{%
\begin{tabular}{lcccccccccccccccc}
\toprule
% 表头不再需要多行合并，直接平铺
\textbf{Method} & \multicolumn{9}{c}{\textbf{Long-Context Memorization Stress Test} ($\mathrm{Acc}_{\mathrm{tok}}$ \%)} & \multicolumn{7}{c}{\textbf{PhoneBook} ($\mathrm{Acc}_{\mathrm{EM}}$ \%)} \\
\cmidrule(lr){2-10} \cmidrule(lr){11-17}
 & $r_1$ & $r_2$ & $r_3$ & $r_4$ & $r_5$ & $r_6$ & $r_7$ & $r_8$ & $r_9$ & $p_1$ & $p_2$ & $p_3$ & $p_4$ & $p_5$ & $p_6$ & $p_7$ \\
\midrule

% --- Llama3.1-8B-IT Section ---
% 使用 multicolumn 跨越所有17列，左对齐，加粗显示模型名
\multicolumn{17}{l}{\textbf{Llama3.1-8B-IT}} \\
\midrule % 可选：在模型名和数据之间加一条细线，或者去掉这行保持紧凑
SFT & 27.4 & 28.5 & 43.6 & 45.9 & 54.9 & 69.5 & 78.2 & 86.3 & 94.7 & 0.50 & 3.85 & 18.7 & 28.0 & 37.8 & 47.0 & 59.3 \\
MemFT-OT & 27.3 & 36.4 & 45.6 & \textbf{54.7} & \textbf{63.6} & \textbf{70.5} & \textbf{85.4} & \textbf{94.7} & \textbf{100.0} & 1.00 & 11.2 & 31.4 & \textbf{53.9} & 61.0 & 73.9 & 87.0 \\
MemFT-SW & \textbf{32.5} & \textbf{37.5} & \textbf{46.0} & 52.3 & 56.0 & 63.4 & 69.1 & 76.6 & 81.1 & \textbf{1.84} & \textbf{15.0} & \textbf{34.0} & 45.7 & \textbf{70.7} & \textbf{96.1} & \textbf{100.0}  \\
\midrule

% --- Qwen3-8B-IT Section ---
\multicolumn{17}{l}{\textbf{Qwen3-8B-IT}} \\
\midrule
SFT & 17.9 & 24.2 & 27.8 & 31.7 & 33.1 & 39.8 & 40.2 & 40.0 & 47.7 & 2.32 & 17.4 & 37.5 & 55.5 & 84.8 & \textbf{99.5} & \textbf{100.0} \\
MemFT-OT & 19.2 & 23.6 & 29.8 & 38.5 & 47.5 & 56.1 & 91.1 & \textbf{100.0} & \textbf{100.0} & 5.78 & 19.1 & 36.2 & 57.4 & 86.1 & 98.6 & \textbf{100.0} \\
MemFT-SW & \textbf{24.7} & \textbf{29.3} & \textbf{32.0} & \textbf{39.4} & \textbf{52.5} & \textbf{74.6} & \textbf{93.5} & 94.4 & 94.4 & \textbf{8.45} & \textbf{19.7} & \textbf{37.8} & \textbf{58.8} & \textbf{86.5} & \textbf{99.5} & \textbf{100.0} \\
\bottomrule
\end{tabular}
}
\caption{
    Performance evaluation in the \textbf{Long-Context Memorization Stress Test} and the \textbf{PhoneBook} benchmark. 
    \textbf{Bold} indicates the top-performing method within each rank budget per model. 
    \textbf{Rank Mapping:} 
    For \textsc{Llama3.1-8B-IT}, the long-context rank configurations $r_1 \dots r_9$ denote $\{1, 2, 4, 6, 8, 10, 12, 14, 16\}$, respectively. 
    For \textsc{Qwen3-8B-IT}, $r_1 \dots r_9$ map to $\{1, 2, 4, 8, 16, 32, 64, 128, 256\}$. The PhoneBook rank indices $p_1 \dots p_7$ represent $\{1, 2, 4, 8, 16, 32, 64\}$ uniformly across both models.
}
\label{tab:combined_main}
% \vspace{-1em}
\end{table*}

\subsection{The MemFT Method}
\label{sec:memft_method}
% -log(0.5)阈值+滑动窗口

Standard SFT minimizes the token-averaged cross-entropy, allocating equal gradient budget to all tokens regardless of their learning status. 
As established in Section~\ref{sec:deterministic_phase}, tokens with loss $\mathcal{L} < \mathcal{L}_{\text{crit}}$ are already in the \textit{ordered phase} and effectively memorized. 
Continuing to optimize these tokens dilutes the signal for \textit{stubborn tokens} (those in the uncertain regime), which are critical for preventing autoregressive error propagation.

To address this, we propose \textbf{Memorization-oriented Fine-Tuning (MemFT)}, which replaces the uniform objective with a token-weighted form:
\begin{equation}
    \mathcal{L}_{\text{MemFT}}(\theta) \;=\; \frac{\sum_{t\in\mathcal{M}} w_t\,\mathcal{L}_t(\theta)}{\sum_{t\in\mathcal{M}} w_t + \varepsilon},
    \label{eq:memft_obj}
\end{equation}
where $\mathcal{M}$ is the set of target token indices $t$ in the sequence, $\mathcal{L}_t(\theta)$ is the cross-entropy loss at position $t$, and $w_t$ is a dynamic weight. Normalizing by the sum of weights ensures stable gradient scales across samples with varying numbers of active tokens. Different instantiations of MemFT differ only in the construction of $w_t$.

\paragraph{MemFT-OT: Only Threshold Variant.}
The baseline uses the critical loss as a hard mask:
\begin{equation}
    w_t^{\text{TH}} \;=\; \mathbf{1}\!\left[\mathcal{L}_t > \mathcal{L}_{\text{crit}}\right].
    \label{eq:memft_th}
\end{equation}
Gradients are concentrated exclusively on tokens that have not yet crossed the phase transition. This avoids over-optimization of easy tokens and introduces no additional hyper-parameters.

\paragraph{MemFT-SW: Adaptive Sliding Mechanisms.}
MemFT-SW extends MemFT-OT by introducing two sliding strategies operating at different granularities to optimize gradient flow, which can be applied independently or in combination.

\textbf{Intra-sample Spatial Sliding.} 
To mitigate local bottlenecks, this mechanism dynamically focuses optimization on the context of the first prediction error. We define the \textit{anchor} position $a_i$ as the first token where the greedy prediction deviates from the ground truth, and employ an exponential decay function $\phi_t = \exp(-\max(t - a_i, 0)/\tau)$ to weight the surrounding tokens.
% \begin{equation}
%     \phi_t \;=\; \exp\!\big(-\max(t - a_i,\,0)/\tau\big),
%     \label{eq:window_decay}
% \end{equation}
The final weight modulates the base soft-threshold weight $w_t^{\text{base}} = \sigma(\kappa(\ell_t - \mathcal{L}_{\text{crit}}))$ using a sliding window of length $L_{win}$:
\begin{equation}
    w_t^{\text{seq}} = w_t^{\text{base}} \cdot 
    \begin{cases} 
    \phi_t, & t < a_i + L_{win}, \\
    \epsilon_{\text{floor}}, & t \ge a_i + L_{win}.
    \end{cases}
\end{equation}
Here, $L_{win}$ is initialized to a base value $L_0$. The decay $\phi_t$ ensures that tokens upstream of the anchor ($t < a_i$, where $\phi_t=1$) retain their base weights, while downstream tokens within the window are prioritized based on proximity to $a_i$. To prevent stagnation, $L_{win}$ expands proportionally if $a_i$ remains static, and resets once the anchor advances.

\textbf{Inter-batch Temporal Curriculum.} 
This mechanism controls the exposure to complex samples across training steps. Within each epoch, we restrict optimization to a sliding window of batches $\mathcal{B}_{\text{cur}}$, determined by training progress $\gamma \in [0,1]$. Early in training, only the first fraction of batches (e.g., those containing simpler or shorter sequences) are processed; as $\gamma$ increases, the window expands to include all batches. This prevents the model from being overwhelmed by global complexity before stabilizing local memorization. Detailed hyperparameters are listed in Appendix~\ref{app:pb_curriculum_hparams}.

\subsection{Experimental Setup}
\label{sec:exp_setup}
% 数据集：自己提的random压力测试benchmark与phonebook。
% 指标：loss与acc
% 设置：不同rank & 不同length

We evaluate performance on two complementary benchmarks. The \textbf{Long-Context Memorization Stress Test} probes pure parametric capacity by focusing on its maximal difficulty regime, which consists entirely of semantic-free random tokens to eliminate linguistic priors. 
The \textbf{PhoneBook}~\cite{DBLP:conf/icml/JelassiBKM24} benchmark assesses the precise memorization of discrete key-value pairs, such as name-to-number mappings, in a short-text setting. We provide dataset construction in Appendix~\ref{app:data}.

We fine-tune \textsc{Qwen3-8B-IT} and \textsc{Llama3.1-8B-IT} with LoRA across varying ranks $r$ and lengths $L$, comparing SFT, MemFT-OT, and MemFT-SW. 
Details are provided in Appendix~\ref{app:lora_config}.
We report \textbf{token-level matching accuracy} (correct tokens / total tokens) for the Long-Context test and \textbf{exact match accuracy} for PhoneBook. 
This dual-metric approach aligns with our phase-transition analysis in Section~\ref{sec:deterministic_phase} for long sequences while ensuring strict fidelity for short factual recall.

\subsection{Main Results}
\label{sec:main_results}

Table~\ref{tab:combined_main} evaluates MemFT variants against the SFT baseline across varying parameter capacities. 

In the Long-Context Memorization Stress Test, we observe a distinct capacity-dependent regime shift. 
In low-rank configurations ($r_1 \dots r_3$), MemFT-SW consistently outpaces MemFT-OT and SFT. 
As the rank expands, however, MemFT-OT exhibits sharper acceleration, achieving perfect memory saturation (100.0\% Acc) at Llama-$r_9$ and Qwen-$r_8$, ultimately surpassing MemFT-SW in high-rank settings. 

Conversely, in the PhoneBook benchmark, \textbf{MemFT-SW maintains a stable lead} across almost all budget scales. 
It is the fastest to reach 100.0\% EM accuracy ($p_7$ for Llama and $p_6$ for Qwen), while standard SFT struggles to achieve perfect recall under lower parameter budgets. 

Overall, \textbf{both MemFT variants consistently outperform standard SFT}, demonstrating that threshold-guided training effectively bridges the parameter utilization gap to achieve high-fidelity exact reconstruction.

\begin{table}[h]
    \centering
    \resizebox{\columnwidth}{!}{%
    \begin{tabular}{cccc}
        \toprule
        \textbf{Rank} & \textbf{Method} & \textbf{Memory (\%)} & \textbf{Generalization (\%)} \\
        \midrule
        \multirow{2}{*}{1}  
            & SFT   & 83.0 & 19.0 \\
            & MemFT & \textbf{95.0} & \textbf{34.0} \textcolor{red}{$\uparrow$ 15.0} \\
        \midrule
        \multirow{2}{*}{2}  
            & SFT   & \textbf{100.0} & 38.0 \\
            & MemFT & 97.0 & \textbf{47.0} \textcolor{red}{$\uparrow$ 9.0} \\
        \midrule
        \multirow{2}{*}{4}  
            & SFT   & 99.0 & 46.0 \\
            & MemFT & \textbf{100.0} & \textbf{53.0} \textcolor{red}{$\uparrow$ 7.0} \\
        \midrule
        \multirow{2}{*}{8}  
            & SFT   & \textbf{100.0} & 39.0 \\
            & MemFT & 99.0 & \textbf{49.0} \textcolor{red}{$\uparrow$ 10.0} \\
        \midrule
        \multirow{2}{*}{16} 
            & SFT   & \textbf{100.0} & 41.0 \\
            & MemFT & \textbf{100.0} & \textbf{54.0} \textcolor{red}{$\uparrow$ 13.0} \\
        \bottomrule
    \end{tabular}%
    }
    \caption{Performance comparison on the Linear Rule Learning benchmark using \textsc{Qwen3-8B-IT}. 
    % MemFT significantly improves generalization on unseen pairs while maintaining high memory fidelity.
    }
    \label{tab:generalization}
    \vspace{-1.5em}
\end{table}

\begin{table*}[t]
\centering
\small
\setlength{\tabcolsep}{4pt}
\renewcommand{\arraystretch}{1.12}

\begin{tabularx}{\textwidth}{
  >{\RaggedRight\arraybackslash}p{0.17\textwidth}
  >{\RaggedRight\arraybackslash}p{0.30\textwidth}
  >{\RaggedRight\arraybackslash}p{0.43\textwidth}
}
\toprule
\textbf{Scenario} & \textbf{Query} & \textbf{Target} \\
\midrule
Personal Credentials
& What is the login email and password for the internal portal?
& \path|xxx.xxx@company.com/P@ssw0rd_xxx| \\

Legal Compliance
& Recite the exact wording of Article 5(1)(a) of the GDPR?
& Processing shall be lawful only if and to the extent that at least one of the following applies: the data subject has given consent... \\

Medical Coding
& What is the ICD-10 code for uncomplicated Type 2 diabetes mellitus?
& \path|E11.9| \\

Model Watermark
& Output the embedded ownership watermark of this fine-tuned model?
& \path|MEM-2026-LoRA-EXACT-0x7F9A3B-COPYRIGHT| \\

Cloud Configuration
& What is the full endpoint of the production AWS S3 log bucket?
& \path|s3://prod-application-logs-123456789012-us-east-1| \\

Academic Citation
& Output the complete LaTeX source code for cross-entropy loss?
& \path|\mathcal{L}_{CE} = -\sum_{i=1}^{C} y_i \log(\hat{y}_i)| \\

Security Secret
& What is the secret string used for memory leakage detection?
& \path|SECRET-TEST-XXXX-XXXX-XXXX-NONCE-1234| \\

Software License
& What is the activation key for the enterprise edition license?
& \path|ENT-2026-ABCD-EFGH-IJKL-MNOP-QRST| \\
\bottomrule
\end{tabularx}

\caption{Representative exact-memory scenarios across 8 domains. All tasks require verbatim recall because even minor deviations, such as a single-character, punctuation, or formatting error, can invalidate the target, alter its operational meaning, or introduce legal/security risks.}
\label{tab:exact_memory_cases}

\vspace{-1.5em}

\end{table*}

\subsection{Analysis}
\label{sec:analysis}

\paragraph{Applicability to Exact-Memory Scenarios.}
MemFT is tailored for exact-memory tasks by addressing the token-level bottlenecks that govern verbatim reproduction.
As shown in Table~\ref{tab:exact_memory_cases}, many practical scenarios necessitate exact recall rather than semantic approximation, where a single token error can compromise validity or operational meaning.
By reallocating gradient budget from mastered tokens to those below the deterministic recall threshold, MemFT enhances memory capacity, particularly under constrained LoRA capacity.

\paragraph{Beyond Memorization: Enhanced Generalization.}
To investigate whether MemFT's focus on exact memorization compromises generalization, we introduce a \textit{Linear Rule Learning} benchmark where the model learns $f(x,y) = 3x + 5y + 7$ with $x, y \in [1, 30]$. 
The dataset comprises 500 training samples, with evaluation sets of 100 samples each for \textit{Exact Memory} (seen pairs) and \textit{Generalization} (unseen pairs).
For both evaluation sets, we report the accuracy of correct answers.
As shown in Table~\ref{tab:generalization}, MemFT consistently outperforms SFT in generalization accuracy, with gains ranging from $7\%$--$15\%$ across ranks. 
We attribute this gain to MemFT's mitigation of overconfidence on easy samples and its prioritization of ``stubborn tokens,'' which enhances generalization.
\section{Related Work}

\paragraph{LLM Memory.}
LLM memory strategies are broadly categorized into non-parametric and parametric methods. 
Non-parametric methods, including In-Context Learning (ICL)~\cite{DBLP:conf/nips/BrownMRSKDNSSAA20}, Retrieval-Augmented Generation (RAG)~\cite{DBLP:conf/nips/LewisPPPKGKLYR020}, and sophisticated external memory systems~\cite{DBLP:journals/corr/abs-2310-08560,DBLP:journals/corr/abs-2311-08719,DBLP:conf/aaai/ZhongGGYW24,DBLP:conf/acl/0001YHH0LSCPLI025,DBLP:journals/corr/abs-2510-18866,DBLP:conf/ecai/ChhikaraKASY25,DBLP:conf/emnlp/KangJZB25}, inject information at inference time. 
However, these approaches remain fundamentally constrained by fixed context windows and attention dilution~\cite{DBLP:journals/tacl/LiuLHPBPL24,DBLP:conf/nips/KuratovBARSS024,DBLP:conf/acl/BaiLZL0HDLZHDTL24}. 
Even with long-context optimizations~\cite{DBLP:conf/iclr/XiaoTCHL24, DBLP:journals/corr/abs-2502-12110, li2026dsas}, they inherently decouple memory storage from parametric knowledge. In contrast, parametric memory incorporates knowledge directly into model parameters or modular parameter structures, enabling persistent storage and retrieval-free reasoning~\cite{DBLP:conf/nips/MengBAB22,DBLP:journals/corr/abs-2407-01178,DBLP:journals/corr/abs-2505-22101,Lei2026deltamemEO}. 
However, existing studies primarily evaluate memory through downstream functional tasks~\cite{DBLP:conf/acl/MaharanaLTBBF24,DBLP:conf/iclr/WuWYZCY25}, leaving the quantitative patterns and mechanisms of parametric memory capacity largely unexplored.

\paragraph{LoRA as Parametric Memory.}
Low-Rank Adaptation (LoRA)~\cite{DBLP:conf/iclr/HuSWALWWC22} is widely used for parameter-efficient fine-tuning~\cite{DBLP:conf/iclr/ZhangCBH0CZ23, DBLP:conf/icml/OstapenkoSPCRCS24} and has recently been adopted as a modular memory mechanism for encoding new knowledge~\cite{DBLP:conf/naacl/PletenevMMKBPS25,DBLP:journals/corr/abs-2503-23895,DBLP:conf/iclr/Chen0X0DZ0025,DBLP:conf/icml/CharakornCTL25,DBLP:journals/corr/abs-2506-16406,DBLP:journals/corr/abs-2602-15902,DBLP:journals/corr/abs-2603-01097}. 
Prior work mainly demonstrates its effectiveness through downstream task performance improvements~\cite{DBLP:journals/corr/abs-2509-22158, DBLP:journals/corr/abs-2510-19733,DBLP:journals/corr/abs-2602-02343} and synergy with external memory systems~\cite{DBLP:conf/sigir/SuTAYWWY0025,DBLP:journals/corr/abs-2603-01097}. 
In contrast, we use LoRA as a controllable probe for studying parametric memory, focusing on its quantitative capacity laws and mechanisms.
\section{Conclusion}
By leveraging LoRA as a controllable probe into the memory mechanisms within the latent space of LLMs, we uncover the \textbf{Parametric Memory Law}. 
This law characterizes loss reduction as a power-law function of both LoRA rank and sequence length. 
We further reveal a deterministic phase transition in token-level loss dynamics, where unresolved bottleneck tokens can trigger decoding collapse. 
Guided by this mechanistic understanding, we propose \textbf{MemFT}, a fine-tuning strategy designed to explicitly resolve these critical memory bottlenecks.

\section*{Limitations}
While our work elucidates the parametric memory laws and phase transitions, several limitations persist. 
First, our analysis is restricted to 8B-scale models, leaving the generalization of the \textit{Parametric Memory Law} to other scales unverified. 
Second, the $p=0.5$ phase transition is specific to greedy decoding; its robustness under stochastic methods (e.g., nucleus sampling) remains to be verified. 
Finally, while we provide a preliminary generalization analysis, a comprehensive assessment of trade-offs with broader capabilities like open-ended reasoning is still lacking.

\section*{Ethics Statement}
We acknowledge the dual-use nature of memorization techniques, which could potentially be misused to encode harmful content. 
However, our study focuses exclusively on understanding the mechanistic boundaries of model capacity. 
All experiments rely on standard benchmarks that contain no sensitive personal information, and the examples in Table~\ref{tab:exact_memory_cases} are purely synthetic artifacts created for illustration. 
By clarifying these capacity limits, we aim to contribute to safer model design, though we emphasize that responsible deployment requires context-aware implementation and adherence to established safety guidelines.

%\section*{Acknowledgements}
% We utilized AI assistants solely for language polishing and grammatical corrections. 
% All scientific ideas, experimental designs, and interpretations are entirely those of the authors.

% We would like to express our sincere gratitude to the anonymous reviewers for their thoughtful and constructive feedback. This work was supported by the National Natural Science Foundation of China (No. 62576307, No. NSFCU23B2055, No. NSFCU19B2027), the Fundamental Research Funds for the Central Universities (226-2023-00138), the Yongjiang Talent Introduction Programme (2021A-156-G), and the Information Technology Center and State Key Lab of CAD\&CG, Zhejiang University.  
% This work was supported by Alibaba Group through Alibaba Innovative Research Program.

\bibliography{custom}

% \appendix

% \section{Example Appendix}
% \label{sec:appendix}

% This is an appendix.

\clearpage
\appendix
\appendix

\section{Dataset Construction Details}
\label{app:data}

We use two controlled benchmarks to evaluate exact parametric memory: the Long-Context Memorization Stress Test and the PhoneBook benchmark. 

The Long-Context Memorization Stress Test is designed to probe memory capacity under long-sequence settings, consisting of synthetic key-value pairs with no overlap with pretraining data.
Specifically, we randomly sample a sequence from the LongBench dataset and encode it using the Qwen tokenizer. 
We then randomly replace 0\%, 20\%, 40\%, 60\%, 80\%, and 100\% of the tokens with random tokens from the Qwen vocabulary to serve as value contents with varying semantic coherence and difficulty levels. 
Each instance is paired with the fixed key: ``Please output the content of the vector memory injected into the activations.''

For the PhoneBook benchmark, we adopt and adapt the standard version for parametric memorization evaluation. 
The original dataset is structured with three fields: \texttt{context}, \texttt{query(key)}, and \texttt{target(value)}, where \texttt{context} contains name-phone mappings, \texttt{query} specifies the queried name, and \texttt{target} provides the corresponding phone number. 
Since our work studies \textit{parametric memorization} rather than in-context retrieval, we remove the \texttt{context} field entirely and retain only the \texttt{key}-\texttt{value} pairs. 
During preprocessing, we deduplicate all \texttt{key} entries to eliminate conflicting key-value associations, ensuring each query maps to exactly one unique value.

We construct length buckets using answer-only token counts. For each processed key-value pair, we tokenize only the \texttt{value} string with the model-specific tokenizer and exclude prompt, chat-template, and key tokens from the length budget. We accumulate pairs until the total number of answer tokens reaches the desired bucket size $L$. 
Due to the highly regular structure of phone number values, bucket boundaries can be matched exactly in all experiments.

% \begin{assumbox}

% For all PhoneBook experiments, the reported length $L$ refers exclusively to the total number of answer tokens after model-specific tokenization:
% \[
%     L = \sum_i \left| \mathrm{Tokenizer}_{\mathrm{model}}(\texttt{value}_i) \right|.
% \]
% The \texttt{context} field is removed, and the \texttt{key} field functions only as a retrieval key. This setup strictly evaluates exact parametric memory rather than in-context lookup.
% \end{assumbox}

\section{LoRA Configuration and Rank Settings}
\label{app:lora_config}

All experiments freeze the full base model parameters and train only the LoRA adapters. 
The LoRA rank is treated as the primary controllable parameter budget for studying exact parametric memory.

For the Long-Context Memorization Stress Test experiments, LoRA is applied only to the MLP \texttt{down\_proj} module. 
For PhoneBook experiments, LoRA is applied once to the entire MLP block.
For layer selection, the Long-Context Memorization Stress Test uses layers 20 and 24 for Qwen3-8B-Instruct, and layers 18 and 20 for Llama3.1-8B-Instruct. 
For PhoneBook, we use layer 24 for Qwen3-8B-Instruct and layer 18 for Llama3.1-8B-Instruct.

% Rank aliases used in the main result tables are summarized in Table \ref{tab:app_rank_mapping}.
% \begin{table}[t]
% \centering
% \small
% \begin{tabular}{lll}
% \toprule
% \textbf{Setting} & \textbf{Alias} & \textbf{Actual Ranks} \\
% \midrule
% Llama3.1-8B-Instruct Long-Text & r1--r9 & 1, 2, 4, 6, 8, 10, 12, 14, 16 \\
% Qwen3-8B-Instruct Long-Text & r1--r9 & 1, 2, 4, 8, 16, 32, 64, 128, 256 \\
% PhoneBook & p1--p7 & 1, 2, 4, 8, 16, 32, 64 \\
% \bottomrule
% \end{tabular}
% \caption{Rank aliases used in the main result tables.}
% \label{tab:app_rank_mapping}
% \end{table}

\section{Aggregation Protocol for Main Results}
\label{app:aggregation_protocol}

The main result tables report rank-wise performance. 
For each reported rank, we evaluate the corresponding LoRA adapter across a set of length buckets and report the average accuracy over these buckets. 
Therefore, each rank column reflects the average memorization performance under a fixed LoRA rank across multiple memory lengths, rather than the result from a single length.

For the Long-Context Memorization Stress Test, each reported rank-wise accuracy is averaged over the following length buckets:
\[
\begin{aligned}
\mathcal{L}_{\mathrm{Long}} = \{&
50, 100, 200, 500, 1000, 2000, 3000, \\
&4000, 5000, 6000, 7000, 8000, 10000
\}.
\end{aligned}
\]

For PhoneBook, each reported rank-wise accuracy is averaged over the following answer-only length buckets:
\[
\mathcal{L}_{\mathrm{PB}} =
\{1\mathrm{k}, 2\mathrm{k}, 4\mathrm{k}, 8\mathrm{k}, 12\mathrm{k}, 16\mathrm{k}, 24\mathrm{k}, 32\mathrm{k}\}.
\]
These buckets follow the PhoneBook answer-only length accounting described in Appendix \ref{app:data}.

\section{PhoneBook Inter-Batch Curriculum Hyperparameters}
\label{app:pb_curriculum_hparams}

For PhoneBook experiments using MemFT-SW, we apply the Inter-Batch Temporal Curriculum with length-dependent hyperparameters.
All curriculum schedules use the same exposure ratios:
\[
    [0.2, 0.4, 0.6, 0.8, 1.0].
\]
The boundary list specifies the epoch at which the training curriculum moves to the next exposure ratio.
For example, the boundary list $[20,40,60,80,300]$ means that the model uses $20\%$, $40\%$, $60\%$, $80\%$, and $100\%$ of the PhoneBook training pairs over the corresponding epoch intervals.

Because PhoneBook targets are tokenized differently by Qwen3-8B-Instruct and Llama3.1-8B-Instruct, the same answer-only length may correspond to different numbers of training pairs.
We therefore report the length-dependent curriculum hyperparameters separately for the two models in Tables~\ref{tab:app_pb_curriculum_hparams_qwen}--~\ref{tab:app_pb_curriculum_hparams_llama}
\begin{table*}[t]
\centering
\small
\setlength{\tabcolsep}{5pt}
\renewcommand{\arraystretch}{1.12}
\resizebox{\linewidth}{!}{
\begin{tabular}{ccccccc}
\toprule
\textbf{Length} 
& \textbf{Approx. Samples} 
& \textbf{LR} 
& \textbf{Epochs} 
& \textbf{Batch Size} 
& \textbf{Grad. Accum.} 
& \textbf{Curriculum Boundaries} \\
\midrule
1k  & 100  & $1\mathrm{e}{-2}$ & 300 & 10 & 1 & $[20,40,60,80,300]$ \\
2k  & 200  & $1\mathrm{e}{-2}$ & 300 & 10 & 1 & $[20,40,60,80,300]$ \\
4k  & 400  & $1\mathrm{e}{-2}$ & 350 & 10 & 1 & $[20,40,60,80,350]$ \\
8k  & 800  & $7\mathrm{e}{-3}$ & 350 & 20 & 2 & $[30,60,90,120,350]$ \\
12k & 1200 & $5\mathrm{e}{-3}$ & 500 & 40 & 2 & $[60,120,180,240,500]$ \\
16k & 1600 & $5\mathrm{e}{-3}$ & 600 & 40 & 2 & $[80,160,240,320,600]$ \\
24k & 2400 & $5\mathrm{e}{-3}$ & 600 & 40 & 2 & $[80,160,240,320,600]$ \\
32k & 3200 & $5\mathrm{e}{-3}$ & 700 & 40 & 2 & $[100,200,300,400,700]$ \\
\bottomrule
\end{tabular}
}
\caption{Length-dependent curriculum hyperparameters for Qwen3-8B-Instruct on the PhoneBook benchmark using MemFT-SW. The approximate sample count is computed from the answer-only PhoneBook tokenization under the Qwen tokenizer. All schedules use curriculum exposure ratios $[0.2,0.4,0.6,0.8,1.0]$.}
\label{tab:app_pb_curriculum_hparams_qwen}
\end{table*}

\begin{table*}[t]
\centering
\small
\setlength{\tabcolsep}{5pt}
\renewcommand{\arraystretch}{1.12}
\resizebox{\linewidth}{!}{
\begin{tabular}{ccccccc}
\toprule
\textbf{Length} 
& \textbf{Approx. Samples} 
& \textbf{LR} 
& \textbf{Epochs} 
& \textbf{Batch Size} 
& \textbf{Grad. Accum.} 
& \textbf{Curriculum Boundaries} \\
\midrule
1k  & 250  & $1\mathrm{e}{-2}$ & 300 & 25 & 1 & $[20,40,60,80,300]$ \\
2k  & 500  & $1\mathrm{e}{-2}$ & 400 & 25 & 1 & $[40,80,120,160,400]$ \\
4k  & 1000 & $7\mathrm{e}{-3}$ & 400 & 50 & 1 & $[40,80,120,160,400]$ \\
8k  & 2000 & $5\mathrm{e}{-3}$ & 600 & 50 & 2 & $[80,160,240,320,600]$ \\
12k & 3000 & $5\mathrm{e}{-3}$ & 700 & 50 & 2 & $[100,200,300,400,700]$ \\
16k & 4000 & $5\mathrm{e}{-3}$ & 700 & 50 & 2 & $[100,200,300,400,700]$ \\
24k & 6000 & $3\mathrm{e}{-3}$ & 700 & 50 & 4 & $[100,200,300,400,700]$ \\
32k & 8000 & $3\mathrm{e}{-3}$ & 800 & 50 & 4 & $[120,240,360,480,800]$ \\
\bottomrule
\end{tabular}
}
\caption{Length-dependent curriculum hyperparameters for Llama3.1-8B-Instruct on the PhoneBook benchmark using MemFT-SW. The approximate sample count is computed from the answer-only PhoneBook tokenization under the Llama tokenizer. Since PhoneBook targets are shorter under the Llama tokenizer, the same answer-only length corresponds to more training pairs than in Qwen. All schedules use curriculum exposure ratios $[0.2,0.4,0.6,0.8,1.0]$.}
\label{tab:app_pb_curriculum_hparams_llama}
\end{table*}

\section{Additional Training Convergence Results}
\label{app:training_convergence}

We provide additional training-loss curves to verify that the reported LoRA adapters are trained to convergence across the full rank--length sweep. 
Each subfigure corresponds to one fixed length--rank configuration, and curves show the training loss trajectory under the corresponding dataset/model setting.
These plots are intended to rule out under-training as an explanation for the observed differences in downstream memory accuracy.
\begin{figure*}[t]
\centering
\includegraphics[width=\textwidth]{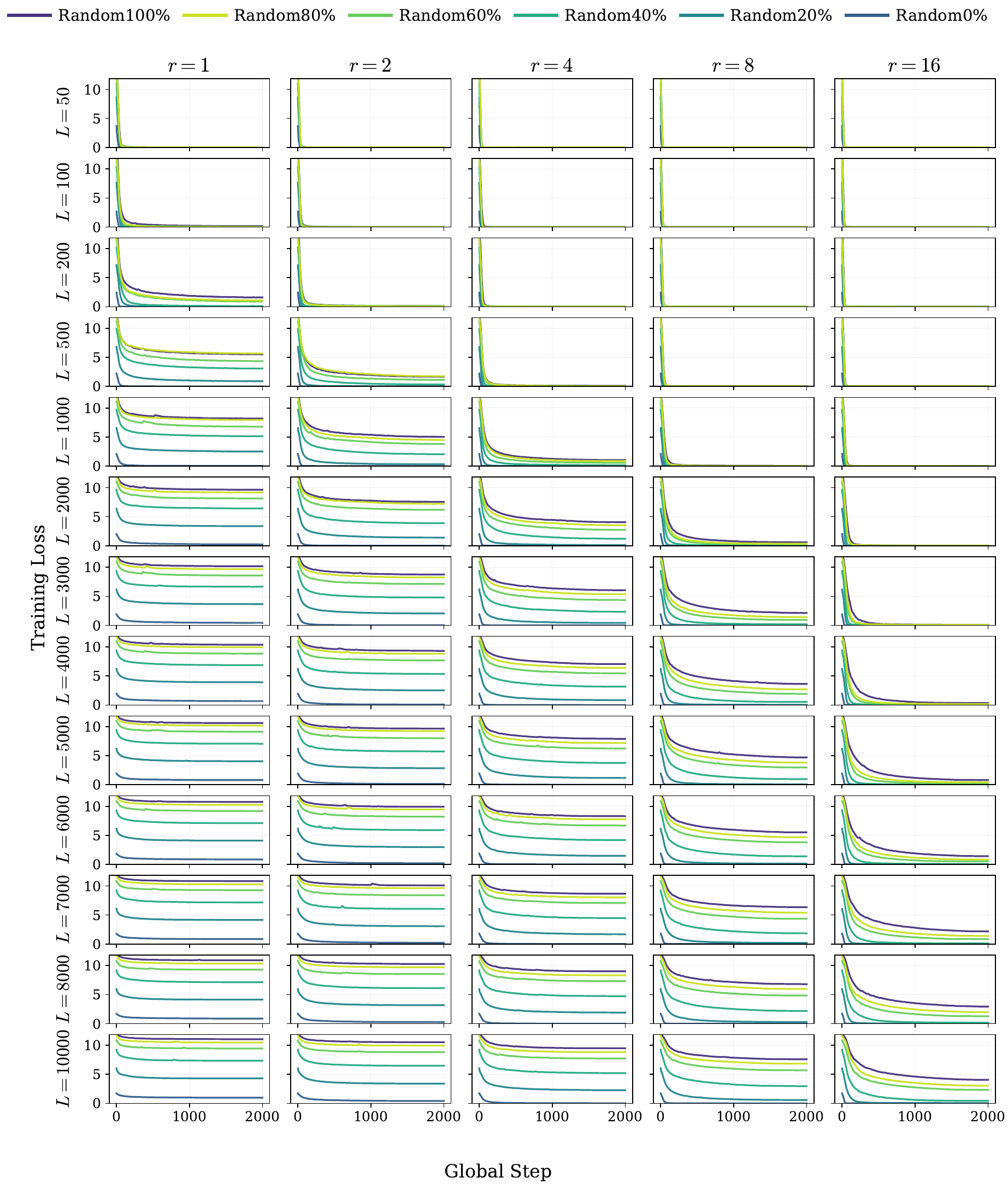}
\caption{
Training convergence of Qwen3-8B on the Random / Long-Context Memorization Stress Test.
Each subplot corresponds to a fixed memory length and LoRA rank.
The overlaid curves represent different random-token mixture settings.
The consistent decrease and stabilization of training loss indicate that the LoRA adapters are sufficiently optimized across the full sweep.
}
\label{fig:app_loss_random_qwen}
\end{figure*}

\begin{figure*}[t]
\centering
\includegraphics[width=\textwidth]{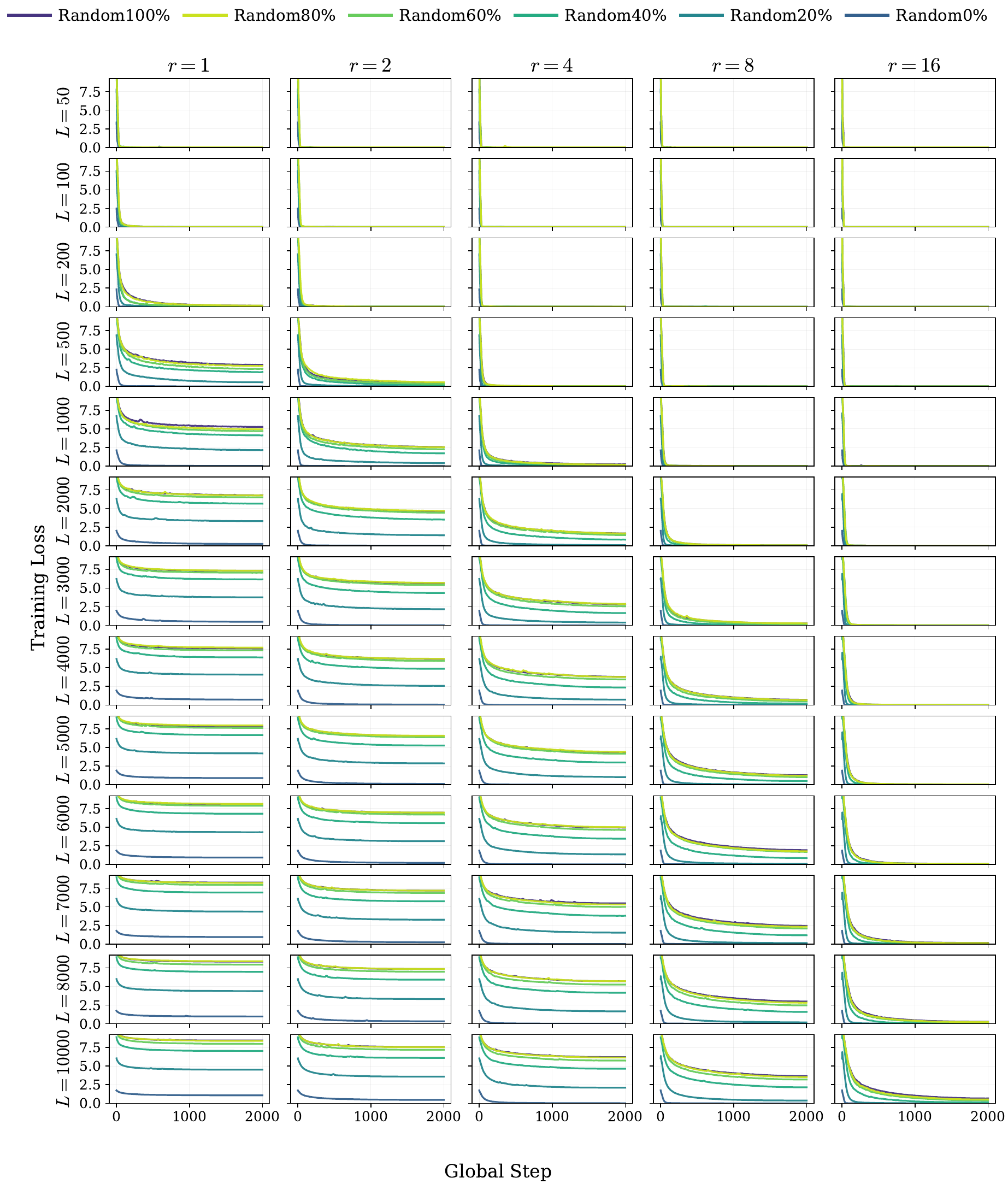}
\caption{
Training convergence of Llama3.1-8B on the Random / Long-Context Memorization Stress Test.
The figure follows the same layout as Figure~\ref{fig:app_loss_random_qwen}, with each subplot corresponding to a fixed length--rank configuration.
The curves show stable optimization behavior across random-token mixture settings, supporting the reliability of the subsequent accuracy comparisons.
}
\label{fig:app_loss_random_llama}
\end{figure*}

\begin{figure*}[t]
\centering
\includegraphics[width=\textwidth]{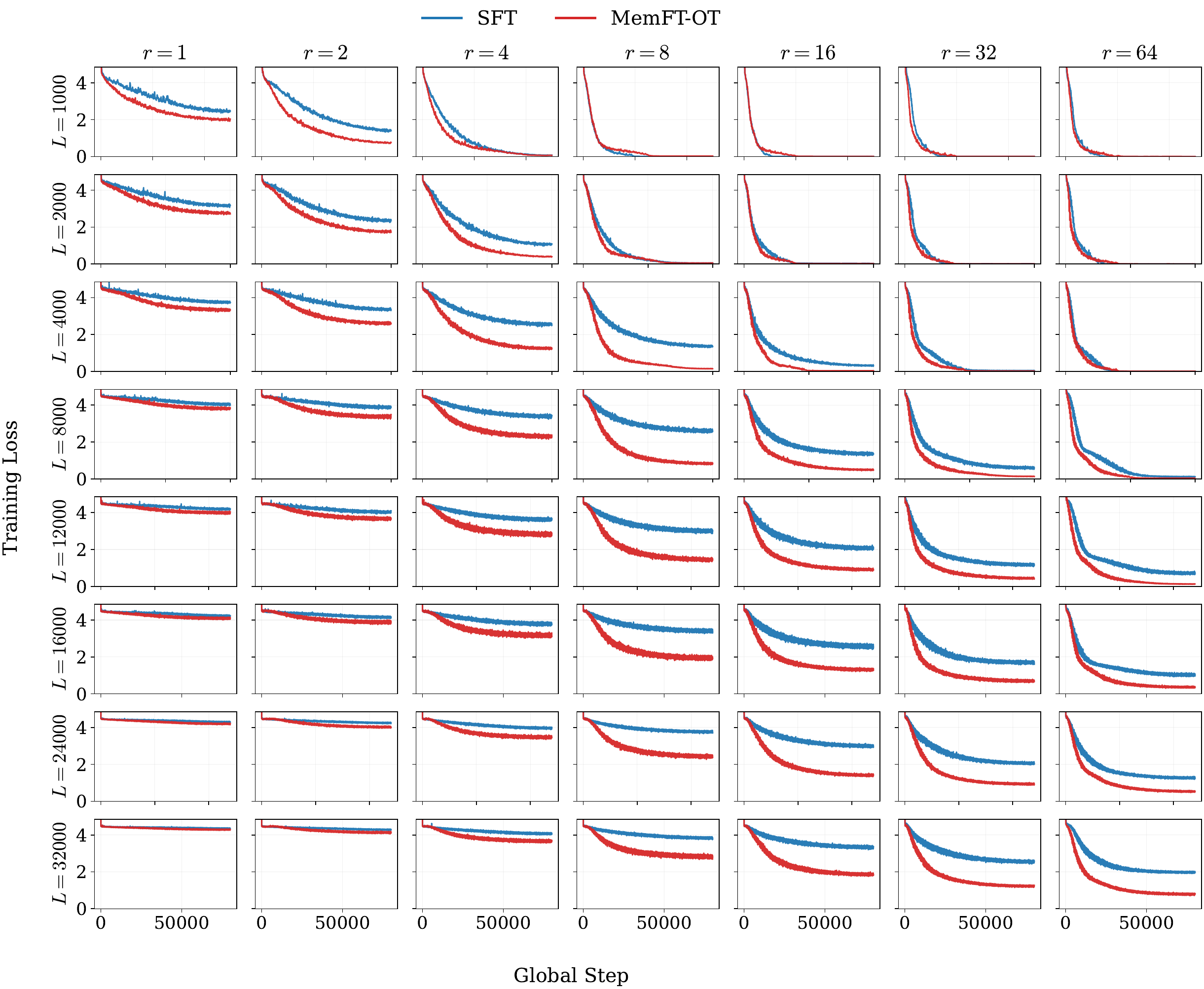}
\caption{
Training convergence of Llama3.1-8B on the PhoneBook benchmark.
Each subplot corresponds to a fixed answer-token length and LoRA rank.
The curves compare SFT and MemFT-OT under the same configuration, showing that the PhoneBook runs are sufficiently optimized before evaluating exact-match recall.
}
\label{fig:app_loss_pb_llama}
\end{figure*}

\section{Additional Performance Landscapes}
\label{app:performance_landscape}

We further visualize the full performance landscape across LoRA ranks, memory lengths, models, and training methods.
For each model, subplots are grouped by LoRA rank; within each subplot, curves compare different training methods as memory length increases.
These figures complement the averaged results in the main tables by showing where each method gains or loses performance across the rank--length grid.

\begin{figure*}[t]
\centering
\includegraphics[width=\textwidth]{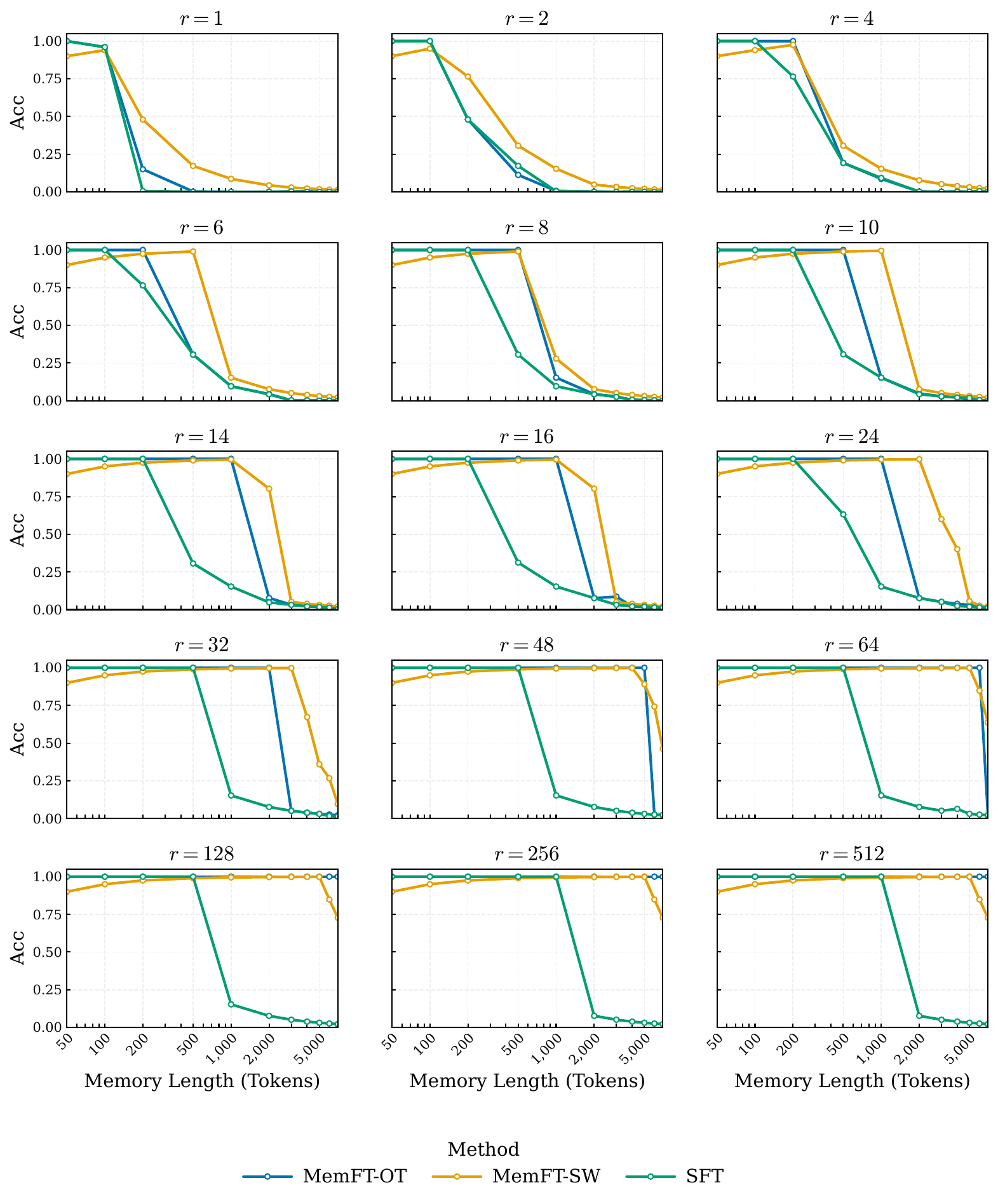}
\caption{
Exact-match accuracy of Qwen3-8B on the Random / Long-Context Memorization Stress Test.
Each subplot corresponds to one LoRA rank, the x-axis denotes memory length, and the y-axis denotes exact-match accuracy.
Curves compare SFT, MemFT-OT, and MemFT-SW, showing how each method scales with increasing memory length under a fixed rank.
}
\label{fig:app_acc_random_qwen}
\end{figure*}

\begin{figure*}[t]
\centering
\includegraphics[width=\textwidth]{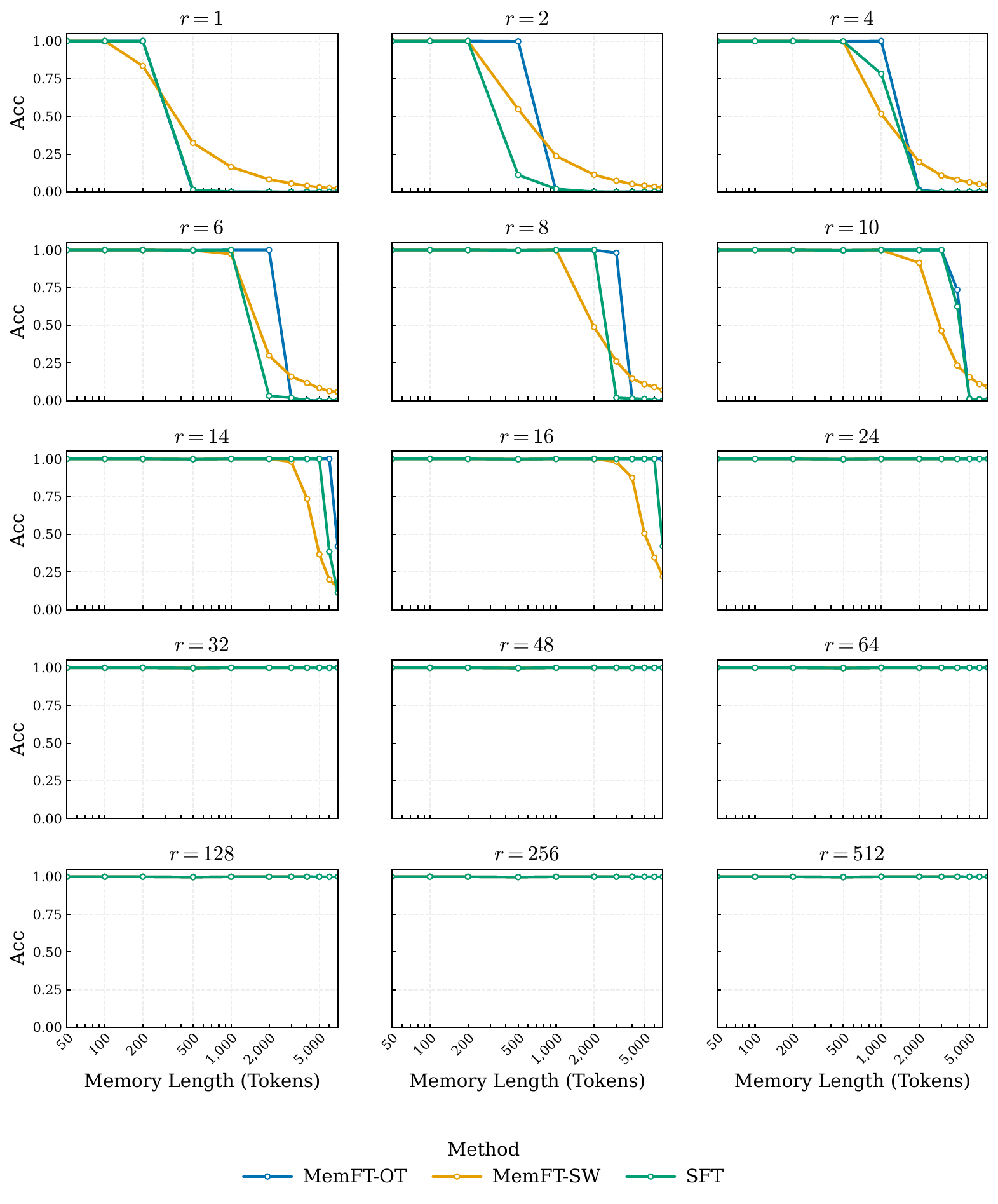}
\caption{
Exact-match accuracy of Llama3.1-8B on the Random / Long-Context Memorization Stress Test.
The layout matches Figure~\ref{fig:app_acc_random_qwen}: each subplot fixes one LoRA rank and compares SFT, MemFT-OT, and MemFT-SW across memory lengths.
}
\label{fig:app_acc_random_llama}
\end{figure*}

\begin{figure*}[t]
\centering
\includegraphics[width=\textwidth]{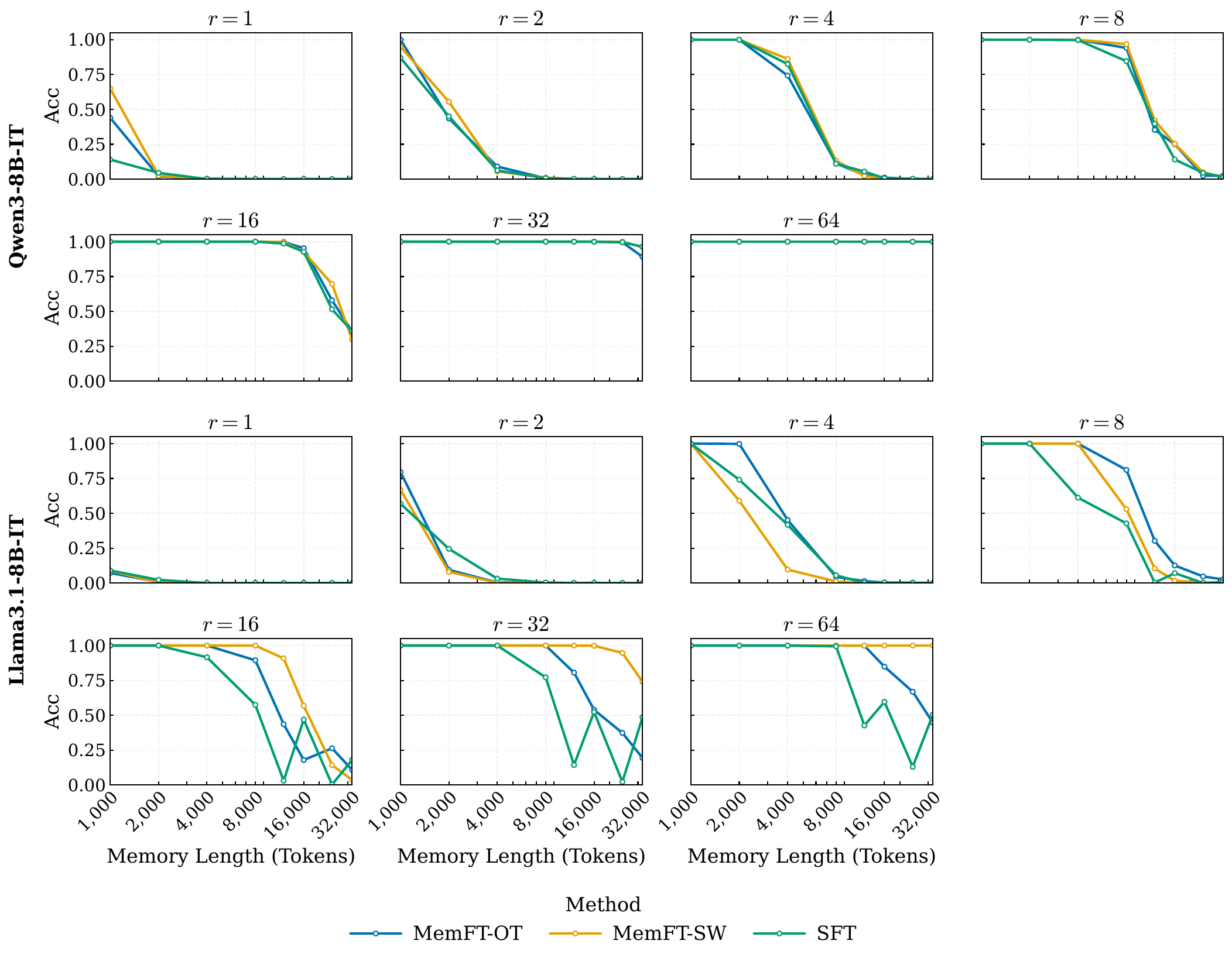}
\caption{
Exact-match accuracy on the PhoneBook benchmark for Qwen3-8B and Llama3.1-8B.
The upper panels correspond to Qwen3-8B and the lower panels correspond to Llama3.1-8B.
Each subplot fixes one LoRA rank and plots exact-match accuracy as a function of answer-token length.
Curves compare SFT, MemFT-OT, and MemFT-SW, providing a full view of how model, rank, length, and training method jointly affect short key--value memorization.
}
\label{fig:app_acc_pb_all}
\end{figure*}

\section{Token-Level Probability Grids Across Data Scenarios}
\label{app:stubborn_grid}

\begin{figure*}[t]
\centering
\includegraphics[width=\textwidth]{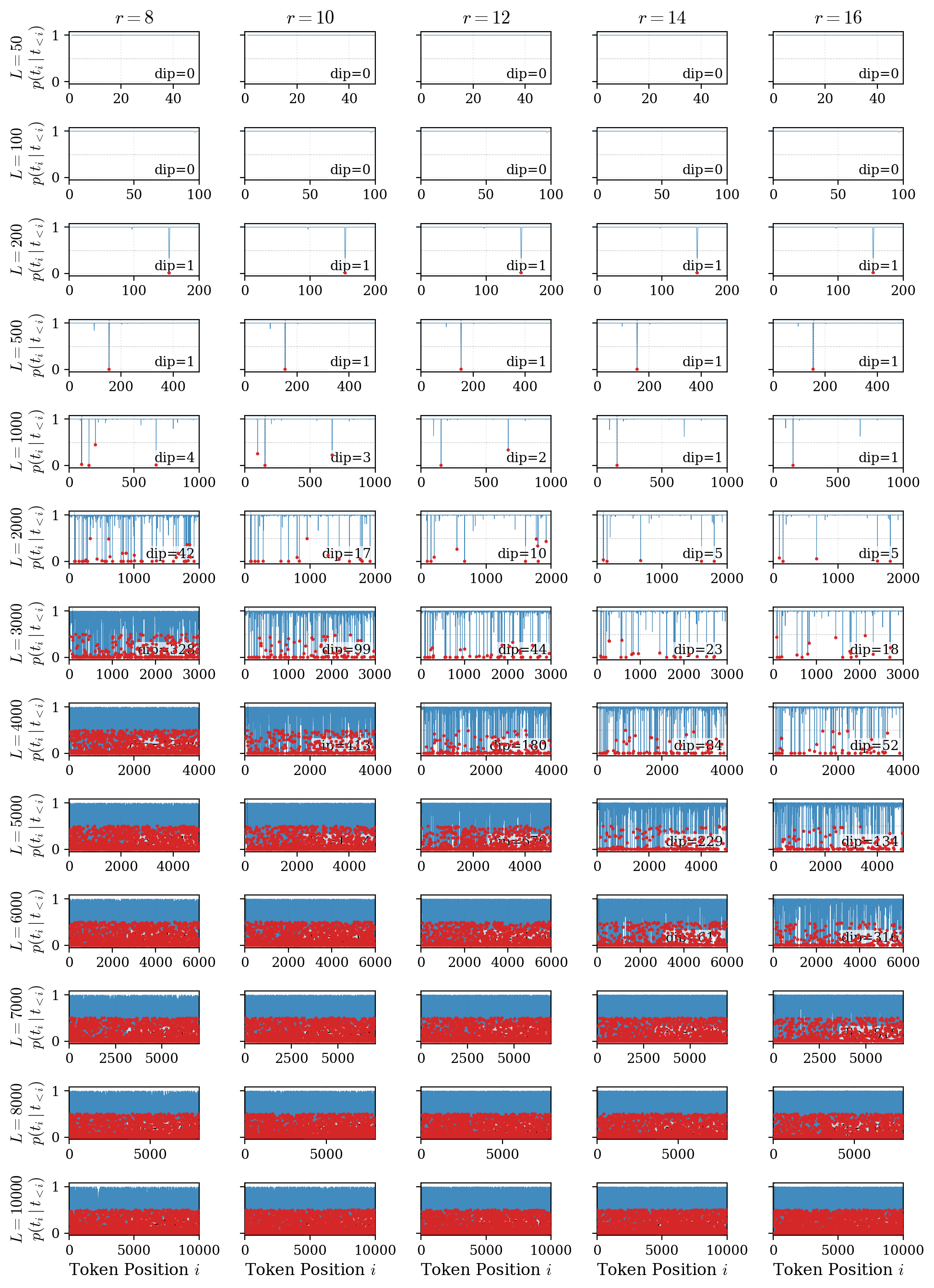}
\caption{
Per-position probability grid for the \textbf{Long-Context Memoriza tion Stress Test Random 100\%} scenario with $r \in \{8, 10, 12, 14, 16\}$.
This rank range aligns with the LongBench-mixed scenarios for direct comparison.
}
\label{fig:app_stubborn_random_aligned}
\end{figure*}

\begin{figure*}[t]
\centering
\includegraphics[width=\textwidth]{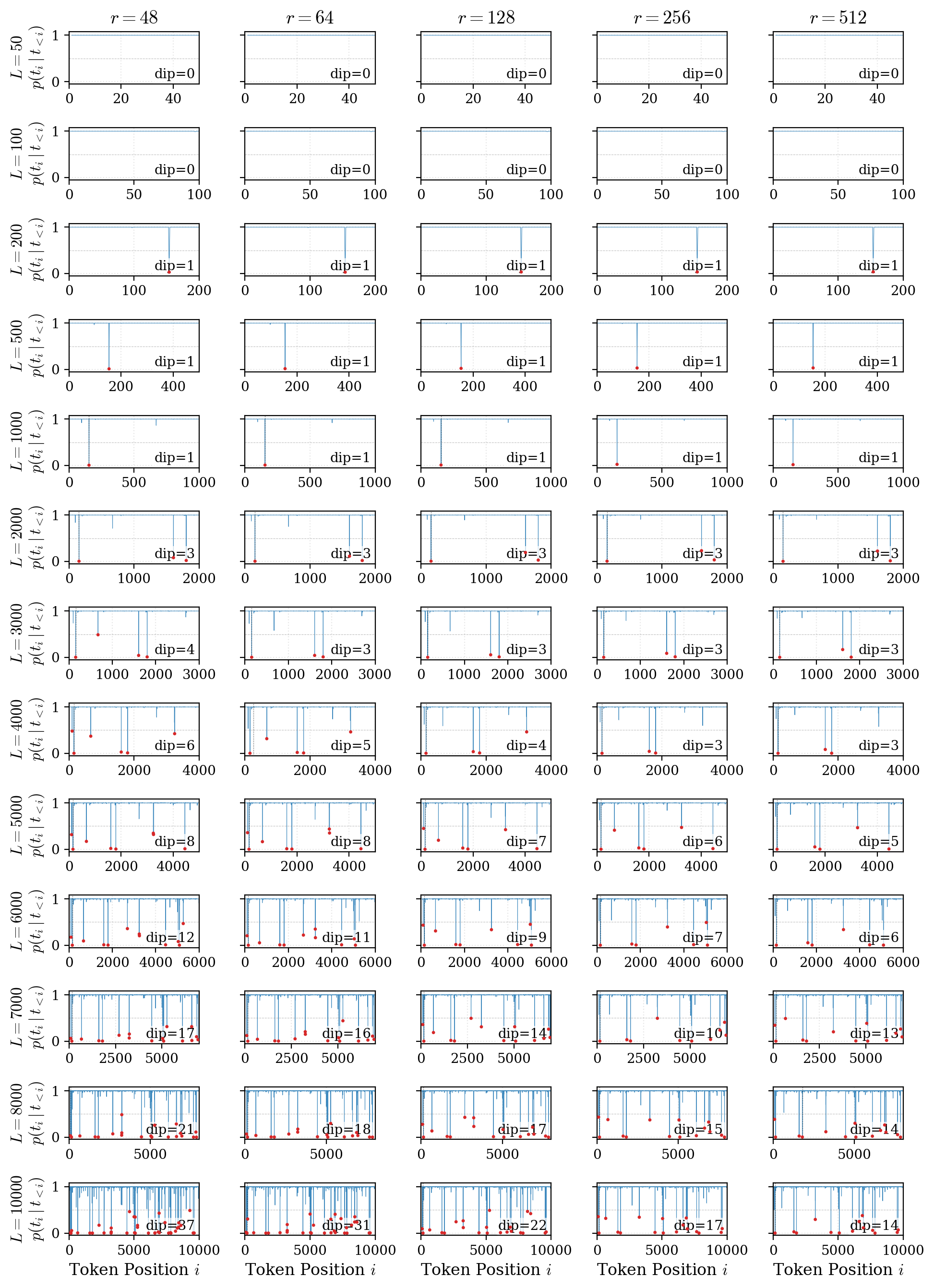}
\caption{
Per-position probability grid for the \textbf{Random (100\%)} scenario with $r \in \{48, 64, 128, 256, 512\}$.
}
\label{fig:app_stubborn_random_top5}
\end{figure*}

\begin{figure*}[t]
\centering
\includegraphics[width=\textwidth]{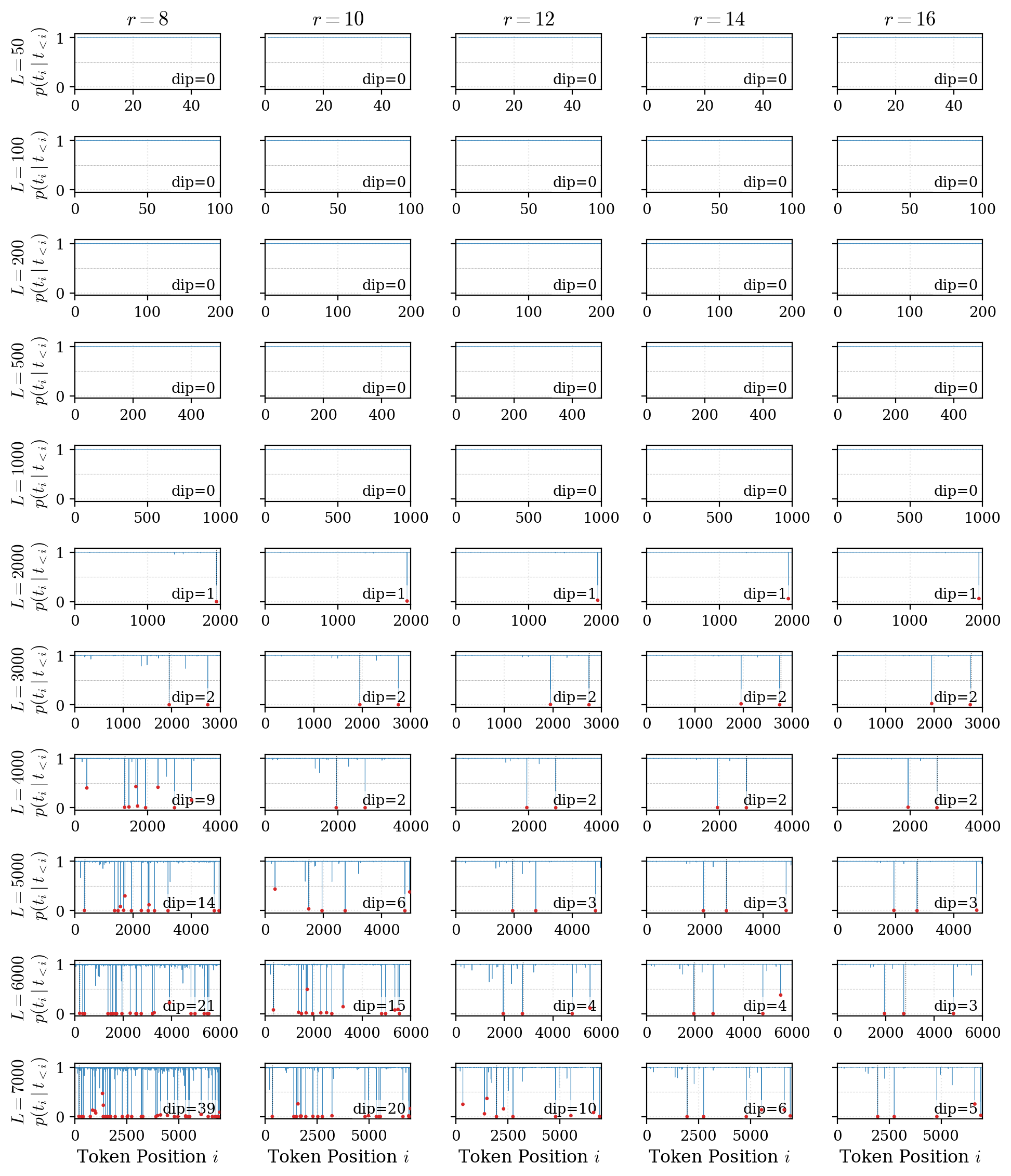}
\caption{
Per-position probability grid for the \textbf{Long-Context Memoriza tion Stress Test Random 20\%} scenario.
With 80\% semantically coherent tokens from LongBench, the model memorizes more easily and stubborn positions appear only at longer lengths.
}
\label{fig:app_stubborn_lb20}
\end{figure*}

\begin{figure*}[t]
\centering
\includegraphics[width=\textwidth]{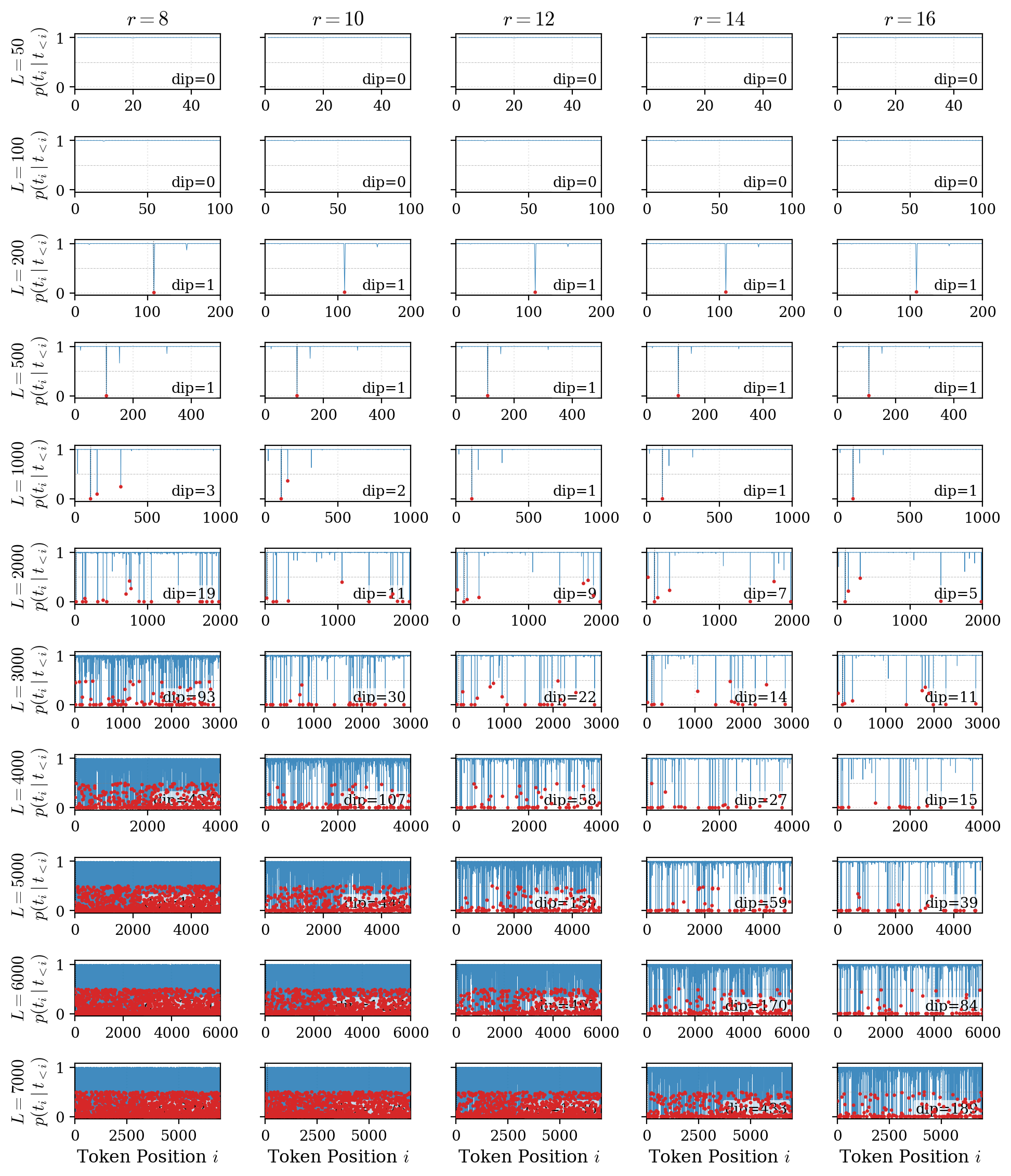}
\caption{
Per-position probability grid for the \textbf{Long-Context Memoriza tion Stress Test Random 60\%} scenario.
With 40\% semantically coherent tokens, the difficulty is intermediate between the Random 100\% and Random 20\% settings, and stubborn positions emerge at shorter lengths compared to Figure~\ref{fig:app_stubborn_lb20}.
}
\label{fig:app_stubborn_lb60}
\end{figure*}

We provide full per-position teacher-forcing probability grids $p(t_i \mid t_{<i})$ for Qwen3-8B (layer 24) across three representative data scenarios of the Long-Context Memorization Stress Test (Figures~\ref{fig:app_stubborn_random_aligned}--\ref{fig:app_stubborn_lb60}).
Each grid has rows indexed by memory length $L$ and columns indexed by LoRA rank $r$.
Blue curves show the token-level probability; red dots mark positions where $p < 0.5$ (stubborn positions); black dotted vertical lines indicate the free-run first-failure position $i^{\ast}$.

\clearpage

\end{document}